%% file: main.tex
\documentclass[journal]{IEEEtran}
\usepackage[hyphens]{url}
\usepackage[pdftex]{graphicx}
\usepackage[latin1]{inputenc}
\usepackage{amssymb,amsmath,array,amsfonts}
\usepackage{color}
\usepackage{booktabs}
\usepackage{subfigure}
\usepackage{multirow}
\usepackage{hyperref}
\usepackage{soul}
\usepackage{ulem}
\usepackage[table,xcdraw]{xcolor}
\usepackage{balance}

\usepackage[style=base]{caption}  
\DeclareCaptionFormat{side}{\hspace*{0.2cm}\parbox{3.9cm}{#1#2#3}}
\DeclareCaptionFormat{side2}{\hspace*{-0.5cm}\parbox{3cm}{#1#2#3}}
\DeclareCaptionFormat{centered}{\hspace*{2.3cm}\parbox{12cm}{#1#2#3}}
\DeclareCaptionFormat{centered2}{\hspace*{0.1cm}\parbox{11.5cm}{#1#2#3}}
\usepackage[rightcaption]{sidecap}
\sidecaptionvpos{figure}{c}

\newcommand{\vect}[1]{\mathbf{#1}}

  

\newcommand{\norm}[2][2]{\left\lVert #2 \right\rVert_{#1}}
\makeatletter
\DeclareRobustCommand{\cev}[1]{%
  \mathpalette\do@cev{#1}%
}
\newcommand{\do@cev}[2]{%
  \fix@cev{#1}{+}%
  \reflectbox{$\m@th#1\vec{\reflectbox{$\fix@cev{#1}{-}\m@th#1#2\fix@cev{#1}{+}$}}$}%
  \fix@cev{#1}{-}%
}
\newcommand{\fix@cev}[2]{%
  \ifx#1\displaystyle
    \mkern#23mu
  \else
    \ifx#1\textstyle
      \mkern#23mu
    \else
      \ifx#1\scriptstyle
        \mkern#22mu
      \else
        \mkern#22mu
      \fi
    \fi
  \fi
}
\DeclareMathOperator*{\argmin}{arg\,min}

\begin{document}

\title{Reservoir computing approaches for representation and classification of multivariate time series}

\author{Filippo Maria Bianchi$^{*}$, 
        Simone Scardapane, 
        Sigurd L{\o}kse,
        and~Robert Jenssen
\thanks{*filippombianchi@gmail.com}
\thanks{F. M. Bianchi is with NORCE, The Norwegian Research Centre}
\thanks{S. Scardapane is with Sapienza University of Rome}
\thanks{S. L{\o}kse and R. Jenssen are with UiT, the Arctic University of Troms{\o}.}
}

\markboth{}%
{Bianchi \MakeLowercase{\textit{et al.}}: Reservoir computing approaches for representation and classification of multivariate time series}

\maketitle

\begin{abstract}
Classification of multivariate time series (MTS) has been tackled with a large variety of methodologies and applied to a wide range of scenarios.
Reservoir Computing (RC) provides efficient tools to generate a vectorial, fixed-size representation of the MTS that can be further processed by standard classifiers. 
Despite their unrivaled training speed, MTS classifiers based on a standard RC architecture fail to achieve the same accuracy of fully trainable neural networks.
In this paper we introduce the \textit{reservoir model space}, an unsupervised approach based on RC to learn vectorial representations of MTS.
Each MTS is encoded within the parameters of a linear model trained to predict a low-dimensional embedding of the reservoir dynamics.
Compared to other RC methods, our model space yields better representations and attains comparable computational performance, thanks to an intermediate dimensionality reduction procedure.
As a second contribution we propose a modular RC framework for MTS classification, with an associated open-source Python library.
The framework provides different modules to seamlessly implement advanced RC architectures.
The architectures are compared to other MTS classifiers, including deep learning models and time series kernels.
Results obtained on benchmark and real-world MTS datasets show that RC classifiers are dramatically faster and, when implemented using our proposed representation, also achieve superior classification accuracy.
\end{abstract}


\section{Introduction}

The problem of classifying multivariate time series (MTS) consists in assigning each MTS to one of a fixed number of classes. 
This is a fundamental task in many applications, including (but not limited to) health monitoring \cite{mikalsen2016learning}, civil engineering \cite{carden2008arma}, action recognition \cite{hunt2015using}, and speech analysis \cite{trentin2015emotion}. 
The problem has been tackled by approaches spanning from the definition of tailored distance measures over MTS to the identification of patterns in the form of dictionaries or shapelets~\cite{salvador2007toward, o2017univariate, bagnall2017great, baydogan2015learning}. 
In this paper we focus on classifiers based on recurrent neural networks (RNNs), which first process sequentially the MTS with a dynamic model, and then exploit the sequence of the model states generated over time to perform classification~\cite{graves2013speechJ}. 

Reservoir computing (RC) is a family of RNN models whose recurrent part is generated randomly and then kept fixed~\cite{lukovsevivcius2009reservoir,scardapane2017randomness}. 
Despite this strong simplification, the recurrent part of the model (the reservoir) provides a rich pool of dynamic features which are suitable for solving a large variety of tasks.
Indeed, RC models achieved excellent performance in time series forecasting~\cite{bianchi2015prediction,7286732}, and process modelling~\cite{rodan2017bidirectional}.
In machine learning, RC techniques were originally introduced under the name echo state networks (ESNs) \cite{jaeger2001echo}; in this paper, we use the two terms interchangeably.

RC-based classifiers represent the MTS either as the last or the mean of all the reservoir states and then process it with a classification algorithm for vectorial data~\cite{ma2016functional, skowronski2006minimum}. 
Despite their unrivaled training speed, these approaches fail to achieve the same accuracy of competing state-of-the-art classifiers~\cite{aswolinskiy2016time}.
To learn more powerful representations, an alternative approach originally proposed in~\cite{chen2013model} and later applied to MTS classification and fault detection~\cite{chen2014learning,aswolinskiy2016time}, advocates to map the inputs in a ``model-based'' feature space where the MTS is represented by the parameters of a model, trained to predict the next input from the current reservoir state.
As a drawback, this approach accounts only for those reservoir dynamics useful to predict the next input and could neglect important information that characterize the MTS.
To overcome this limitation, we propose a new model-space criterion that disentangles from the constraints imposed by this formulation.

\subsubsection*{Contributions of the paper}

We propose an unsupervised procedure to generate MTS representations, called \textit{reservoir model space}, that consists in the parameters of the one-step-ahead predictor that estimates the future \textit{reservoir} state, as opposed to the future MTS input. 
As shown in our previous work~\cite{lokse2017training}, the reservoir states carry all the information necessary to reconstruct the phase space that, in turn, gives a complete knowledge of the underlying dynamical system generating the observed MTS.
Therefore, a model capable of predicting the next reservoir state accounts for all the system dynamics and provides a much more accurate characterization of the MTS.
Due to the large size of the reservoir, a na{\"i}ve formulation of the model space yields extremely large representations that lead to overfit in the subsequent classifier and hamper the computational efficiency proper of the RC paradigm.
We address this issue by training the prediction model on a low-dimensional embedding of the original dynamics. 
The embedding is obtained by applying to the reservoir states sequence a modified version of principal component analysis (PCA) for tensors, which keeps separated the modes of variation among time steps and data samples.
The proposed representation is novel and, while our focus is on MTS, it naturally extends also to univariate time series.

As a second contribution, we introduce a unified RC framework (with an associated open source Python library) for MTS classification that generalizes both classic and advanced RC architectures.
Our framework consists of four independent modules that specify i) the architecture of the reservoir, ii) a dimensionality reduction procedure applied to reservoir activations, iii) the representation used to describe the input MTS, and iv) the readout that performs the final classification.

In the experiments, we compare several RC architectures implemented with our framework, with state-of-the-art time series classifiers, classifiers based on fully trainable RNNs, deep learning models, DTW, and SVM configured with kernels for MTS.
The results obtained on several real-world dataset show that the RC classifiers are dramatically faster than the other methods and, when implemented using our proposed representation, also achieve a competitive classification accuracy.
\subsubsection*{Notation}
we denote variables as lowercase letters ($x$); constants as uppercase letters ($X$); vectors as boldface lowercase letters ($\vect{x}$); matrices as boldface uppercase letters ($\vect{X}$); tensors as calligraphic letters ($\mathcal{X}$). 
All vectors are assumed to be columns. The operator $\norm[p]{\cdot}$ is the standard $\ell_p$ norm in Euclidean spaces.
The notation $x(t)$ indicates time step $t$ and $x[n]$ sample $n$ in the dataset.
%
\section{Preliminaries}
\label{sec:rnn}
We consider classification of generic $F$-dimensional MTS with $T$ time instants, whose observation at time $t$ is denoted as $\mathbf{x}(t) \in \mathbb{R}^F$. We represent a MTS in a compact form as a $T \times F$ matrix $\vect{X} = \left[ \mathbf{x}(1), \ldots, \mathbf{x}(T) \right]^T$ 
\footnote{Since MTS may have different lengths, $T$ is a function of the MTS.}.

Common in machine learning is to express the classifier as a combination of an \textit{encoding} and a \textit{decoding} function.
The encoder generates a representation of the input, while the decoder is a discriminative (or predictive) model that computes the posterior probability of the output given the encoder representation. 
An encoder based on an RNN~\cite{bianchi2017recurrent} is particularly suitable to model sequential data, and is governed by the state-update equation
\begin{equation}
\label{eq:general_RNN_state}
    \mathbf{h}(t) = f \left(\mathbf{x}(t), \mathbf{h}(t-1); \theta_{\text{enc}} \right),
\end{equation}
where $\mathbf{h}(t)$ is the RNN state at time $t$ that depends on its previous value $\mathbf{h}(t-1)$ and the current input $\mathbf{x}(t)$, $f(\cdot)$ is a nonlinear activation function (e.g., a sigmoid or hyperbolic tangent), and $\theta_{\text{enc}}$ are adaptable parameters. 
The simplest (vanilla) formulation reads:
\begin{equation}
\label{eq:vanilla_RNN_state}
    \mathbf{h}(t) = tanh \left( \mathbf{W}_\text{in} \mathbf{x}(t) + \mathbf{W}_\text{r} \mathbf{h}(t-1) \right),
\end{equation}
with $\theta_{\text{enc}} = \left\{\mathbf{W}_\text{in}, \mathbf{W}_\text{r} \right\}$. 
The matrices $\mathbf{W}_\text{in}$ and $\mathbf{W}_\text{r}$ are the weights of the input and recurrent connections, respectively.

From the sequence of the RNN states generated over time, $\mathbf{H} = \left[ \mathbf{h}(1), \ldots, \mathbf{h}(T) \right]^T$, it is possible to extract a representation $\mathbf{r}_\vect{X} = r(\mathbf{H})$ of the input $\vect{X}$. 
A common choice is to take $\mathbf{r}_\vect{X} = \mathbf{h}(T)$, since the RNN can embed into its last state all the information required to reconstruct the original input~\cite{sutskever2014sequence}.
The decoder maps the MTS representation $\mathbf{r}_\mathbf{X}$ into the output space, which are the class labels $\mathbf{y}$ for a classification task:
\begin{equation}
    \label{eq:rnn_output}
    \mathbf{y} = g(\mathbf{r}_\mathbf{X}; \theta_\text{dec}) \,,
\end{equation}
where $g(\cdot)$ can be a (feed-forward) neural network or a linear model, and $\theta_\text{dec}$ are the trainable parameters. 

In the following, we describe two RNN-based approaches for MTS classification.
The first is based on fully trainable architectures, the second on RC where the RNN encoder is left untrained.

\subsection{Fully trainable RNNs and gated architectures}
\label{sec:gated_rnns}

In fully trainable RNNs, given a set of MTS $\left\{\vect{X}[n]\right\}_{n=1}^N$ and associated labels $\left\{\vect{y}[n]\right\}_{n=1}^N$, the encoder parameters $\theta_{\text{enc}}$ and the decoder parameters $\theta_{\text{dec}}$ are jointly learned by minimizing an empirical cost:
\begin{equation}
\theta_{\text{enc}}^*, \theta_\text{dec}^* = \underset{\theta_{\text{enc}}, \theta_\text{dec}}{\arg\min} \,\, \frac{1}{N} \sum_{n=1}^N l\bigg(\vect{y}[n], g\Big(r\big(f(\vect{X}[n])\big)\Big)\bigg) \,,
\label{eq:loss_function}
\end{equation}
where $l(\cdot, \cdot)$ is a generic loss function (e.g., cross-entropy over the labels). 
The gradient of \eqref{eq:loss_function} with respect to $\theta_{\text{enc}}$ and $\theta_\text{dec}$ can be computed by back-propagation through time \cite{graves2013speechJ}.

The parameters in the encoding and decoding functions are commonly regularized with a $\ell_2$ norm penalty, controlled by a scalar $\lambda$.
It is also possible to include a dropout regularization, that randomly drops connections during training with probability $p_\text{drop}$~\cite{srivastava2014dropout}. 
In our experiments, we apply a dropout specific for recurrent architectures~\cite{zaremba2014recurrent}.

Despite the theoretical capability of basic RNNs to model any dynamical system, in practice their effectiveness is hampered by the difficulty of training their parameters~\cite{pascanu2013difficulty}.
To ensure stability, the derivative of the recurrent function in an RNN must not exceed unity.
However, as an undesired effect, the gradient of the loss shrinks when back-propagated in time through the network.
Using RC models (described in the next section) is one way of avoiding this problem. 
Another solution is using the long short-term memory (LSTM) network~\cite{hochreiter1997long}, which exploits gating mechanisms to maintain its internal memory unaltered for long time intervals.
However, LSTM flexibility comes at the cost of a higher computational and architectural complexity.
A popular variant is the gated recurrent unit (GRU)~\cite{cho2014learning}, that 
provides a better memory conservation by using less parameters than LSTM.

\subsection{Reservoir computing and output model space}
\label{sec:esn}

To avoid the costly operation of back-propagating through time, the RC approach takes a radical different direction: it still implements the encoding function in \eqref{eq:vanilla_RNN_state}, but the encoder parameters $\theta_\text{enc} = \{\mathbf{W}_\text{in}$, $\mathbf{W}_\text{r} \}$ are randomly generated and left untrained.
To compensate for this lack of adaptability, a large recurrent layer, the \textit{reservoir}, generates a rich pool of heterogeneous dynamics useful to solve many different tasks.
The generalization capabilities of the reservoir mainly depend on three ingredients: (i) a high number of processing units in the recurrent layer, (ii) sparsity of the recurrent connections, and (iii) a spectral radius of the connection weights matrix $\mathbf{W}_\text{r}$, set to bring the system to the edge of stability~\cite{7765110}.
The behaviour of the reservoir is controlled by modifying the following hyperparameters:
the spectral radius $\rho$; the percentage of non-zero connections $\beta$; and the number of hidden units $R$.
Another important hyperparameter is the scaling $\omega$ of the values in $\mathbf{W}_\text{in}$, which controls the amount of nonlinearity in the processing units and, jointly with $\rho$, can shift the internal dynamics from a chaotic to a contractive regime~\cite{livi2018determination}.
A Gaussian noise with standard deviation $\xi$ can also be added in the state update function \eqref{eq:vanilla_RNN_state} for regularization purposes~\cite{jaeger2001echo}.

In ESNs, the decoder (commonly referred as readout) is usually a linear model:
\begin{equation}
    \label{eq:esn_class}
    \mathbf{y} = g(\mathbf{r}_\vect{X}) = \mathbf{V}_o \mathbf{r}_\mathbf{X} + \mathbf{v}_o
\end{equation}
The decoder parameters $\theta_\text{dec} = \{\mathbf{V}_o$, $\mathbf{v}_o \}$ can be learned by minimizing a ridge regression loss function
\begin{equation}
    \label{eq:rdige}
    \theta_\text{dec}^{*} = \argmin \limits_{\{\mathbf{V}_o, \mathbf{v}_o \}} \frac{1}{2} \left\| \mathbf{r}_\mathbf{X}\mathbf{V}_o + \mathbf{v}_o - \mathbf{y} \right\|^2 + \lambda \left\| \mathbf{V}_o \right\|^2,
\end{equation}
which admits a closed-form solution~\cite{scardapane2017randomness}.
The combination of an untrained reservoir and a linear readout defines the basic ESN model~\cite{jaeger2001echo}. 

A powerful representation $\mathbf{r}_\mathbf{X}$ is the \textit{output model space}~\cite{chen2013model, gong2018multiobjective, wang2016effective}, obtained by first processing each MTS with the same reservoir and then training a ridge regression model to predict the input one step-ahead:
\begin{equation}
    \label{eq:esn_pred}
    \mathbf{x}(t+1) = \mathbf{U}_o \mathbf{h}(t) + \mathbf{u}_o.
\end{equation}
The parameters $\mathbf{\theta}_o = \left[ \mathrm{vec}(\mathbf{U}_o); \mathbf{u}_o \right] \in \mathbb{R}^{F(R+1)}$ becomes the representation $\vect{r}_\vect{X}$ of the MTS, which is, in turn, processed by the classifier in \eqref{eq:esn_class}.
In the following, we propose a new model space that yields a more expressive representation of the input.

\section{Proposed reservoir model space representation}
\label{sec:reservoir_space}

In this section we introduce the main contribution of this paper, the \textit{reservoir model space} for representing a (multivariate) time series, and a dimensionality reduction method that extends PCA to multidimensional temporal data.
Related to our idea, but framed in a setting different from RC, are the recent deep learning architectures that learn unsupervised representations by predicting the future in a small-dimensional latent space with autoregressive models~\cite{oord2018representation}.

\subsection{Formulation of the reservoir model space}

The generalization capability of the reservoir is grounded on the large amount of heterogeneous dynamics it generates from the input. 
To predict the next input values, different dynamics are selected depending on the forecast horizon of interest. 
Therefore, when fixing the prediction step (e.g., 1 step-ahead) all those dynamics that are not useful to solve the task are discarded. 
This introduces a bias in the output model space, since the features that are not important for the prediction task can still be useful to characterize the MTS. 
Therefore, we propose a new model space, where each MTS is represented by the parameters of a linear model, which predicts the next reservoir state by accounting for all the reservoir dynamics.
The linear model trained to predict the next reservoir state reads
\begin{equation}
    \label{eq:res_pred}
    \mathbf{h}(t+1) = \mathbf{U}_h \mathbf{h}(t) + \mathbf{u}_h,
\end{equation}
and $\vect{r}_{\vect{X}} = \mathbf{\theta}_h = \left[ \mathrm{vec}(\mathbf{U}_h); \mathbf{u}_h \right] \in \mathbb{R}^{R(R+1)}$ is our proposed representation.

The reservoir model space representation characterizes a generative model of the reservoir sequence
\begin{equation}
    \label{eq:gen_model}
    p\left( \mathbf{h}(T), \mathbf{h}(T-1), \dots, \mathbf{h}(1); \vect{r}_\vect{X} \right).
\end{equation}
The model \eqref{eq:gen_model} provides a characterization of both the input and the generative process of its high-level dynamical features, and also induces a metric relationship between samples~\cite{ng2002discriminative}. 
A classifier that processes the reservoir model representation combines the explanatory capability of generative models with the classification power of the discriminative methods.

\subsection{Dimensionality reduction for reservoir states tensor}

Due to the high dimensionality of the reservoir, the number of parameters of the prediction model in \eqref{eq:res_pred} would grow too large, making the proposed representation intractable.
Drawbacks in using large representations include overfitting and the high amount of computational resources to evaluate the ridge regression solution for each MTS.
In the context of RC, applying PCA to reduce dimensionality of the last reservoir state has shown to improve performance achieved on the inference task~\cite{esn_bidir}.
Compared to non-linear methods for dimensionality reduction such as kernel-PCA or autoencoders~\cite{BIANCHI2019106973}, PCA provides competitive generalization capabilities when combined with RC models and can be computed quickly, thanks to its linear formulation~\cite{lokse2017training}.

Our proposed MTS representation do not coincides with the last reservoir state, but depends on the whole sequence of states generated over time. 
Therefore, we conveniently describe our dataset as a 3-mode tensor $\boldsymbol{\mathcal{H}} \in \mathbb{R}^{N \times T \times R}$ and require a transformation to map $R \rightarrow D$ s.t. $D \ll R$, while maintaining the other dimensions unaltered.
Dimensionality reduction on high-order tensors can be achieved through Tucker decomposition, which decomposes a tensor into a core tensor (the lower-dimensional representation) multiplied by a matrix along each mode.
When only one dimension of $\boldsymbol{\mathcal{H}}$ is modified, Tucker decomposition becomes equivalent to applying a two-dimensional PCA on a specific matricization of $\boldsymbol{\mathcal{H}}$~\cite{kolda2009tensor}. 
Specifically, to reduce the third dimension ($R$) one computes the mode-3 matricization of $\boldsymbol{\mathcal{H}}$ by arranging the mode-3 fibers (high-order analogue of matrix rows/columns) to be the rows of a resulting matrix $\mathbf{H}_{(3)} \in \mathbb{R}^{NT \times R}$. 
Then, standard PCA projects the rows of $\mathbf{H}_{(3)}$ on the eigenvectors associated to the $D$ largest eigenvalues of the covariance matrix $\mathbf{C} \in \mathbb{R}^{R \times R}$, defined as
\begin{equation}
    \label{eq:pca}
    \mathbf{C} = \frac{1}{NT-1}\sum \limits_{i=1}^{NT} \left( \mathbf{h}_i - \mathbf{\bar{h}} \right) \left( \mathbf{h}_i - \mathbf{\bar{h}} \right)^T.
\end{equation}

In \eqref{eq:pca}, $\mathbf{h}_i$ is the $i$-th row of $\mathbf{H}_{(3)}$ and $\mathbf{\bar{h}} = \frac{1}{N} \sum_i^{NT} \mathbf{h}_i$.
As a result of the concatenation of the first two dimensions in $\boldsymbol{\mathcal{H}}$, $\mathbf{C}$ evaluates the variation of the components in the reservoir states across all samples and time steps at the same time. 
Consequently, both the original structure of the dataset and the temporal orderings are lost, as the reservoir states relative to different samples and generated in different time steps are mixed together.
This may lead to a potential loss in the representation capability, as the existence of modes of variation in time courses within individual samples is ignored.

To address this issue, we consider as individual samples the matrices $\mathbf{H}_n \in \mathbb{R}^{T \times R}$, obtained by slicing $\boldsymbol{\mathcal{H}}$ across its first dimension ($N$).
Our proposed sample covariance matrix reads
\begin{equation}
    \label{eq:pca2}
    \mathbf{S} = \frac{1}{N-1} \sum \limits_{n=1}^{N} \left( \mathbf{H}_n - \mathbf{\bar{H}} \right)^T \left( \mathbf{H}_n - \mathbf{\bar{H}} \right).
\end{equation}
The first $D$ leading eigenvectors of $\mathbf{S}$ are stacked in a matrix $\mathbf{E} \in \mathbb{R}^{R \times D}$ and the desired tensor of reduced dimensionality is obtained as $\boldsymbol{\mathcal{\hat{H}}} = \boldsymbol{\mathcal{H}} \times_3 \mathbf{E}$, where, $\times_3$ denotes the 3-mode product. 
Like $\mathbf{C}$, $\mathbf{S} \in \mathbb{R}^{R \times R}$ describes the variations of the variables in the reservoir.
However, since the whole sequence of reservoir states is treated as a single observation, the temporal ordering in different MTS is preserved.

\begin{figure}
	\centering
	\includegraphics[keepaspectratio,width=\columnwidth]{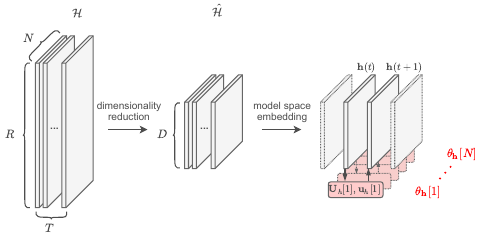}	
    \caption{Schematic depiction of the procedure to generate the reservoir model space representation. 
    For each input MTS $\mathbf{X}[n]$ a sequence of states $\vect{H}[n]$ is generated by a fixed reservoir. 
    Those are the frontal slices (dimension $N$) of $\boldsymbol{\mathcal{H}}$, but notice that in the figure lateral slices (dimension $T$) are shown. 
    The proposed dimensionality reduction reduces the reservoir features from $R$ to $D$. 
    An independent model is trained to predict  $\mathbf{\hat{H}}[n]$, the $n$-th frontal slice of $\boldsymbol{\mathcal{\hat{H}}}$, and its parameters $\mathbf{\theta}_h[n]$ become the representation of $\mathbf{X}[n]$.}
	\label{fig:rmesn_schema}
\end{figure}

After dimensionality reduction, the model in \eqref{eq:esn_pred} becomes
\begin{equation}
    \label{eq:reservoir_pred}
    \mathbf{\hat{h}}(t+1) = \mathbf{U}_h \mathbf{\hat{h}}(t) + \mathbf{u}_h,
\end{equation}
where $\mathbf{\hat{h}}(\cdot)$ are the columns of a frontal slice $\mathbf{\hat{H}}$ of $\boldsymbol{\mathcal{\hat{H}}}$, $\mathbf{U}_h \in \mathbb{R}^{D \times D}$, and $\mathbf{u}_h \in \mathbb{R}^{D}$.
The representation will now coincide with the parameters vector $\mathbf{r}_\mathbf{X} = \mathbf{\theta}_h = \left[ \mathrm{vec}(\mathbf{U}_h); \mathbf{u}_h \right] \in \mathbb{R}^{D(D+1)}$, as shown in Fig.~\ref{fig:rmesn_schema}.

The complexity for computing the reservoir model space representations for all the MTS in the dataset is given by the sum of $\mathcal{O}(NTVH)$, the cost for computing all the reservoir states, and $\mathcal{O}(H^2NT + H^3)$, the cost of the dimensionality reduction procedure.

\section{A unified reservoir computing framework for time series classification}
\label{sec:framework}

\begin{figure*}
	\centering
	\includegraphics[keepaspectratio,width=0.8\textwidth]{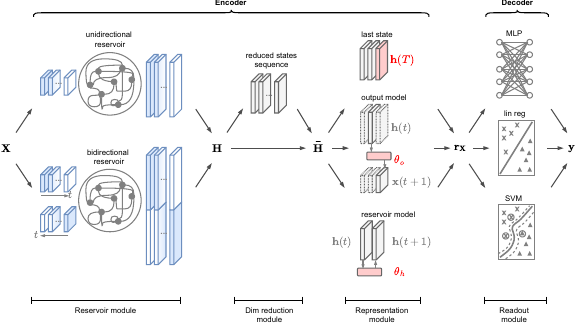}	
    \caption{Framework overview. The encoder generates a representation $\mathbf{r}_\mathbf{X}$ of the MTS $\mathbf{X}$, while the decoder predicts the label $\mathbf{y}$. Several models are obtained by selecting one variant for each module. Arrows indicate mutually exclusive choices}.
	\label{fig:main_schema}
\end{figure*}

In the last years, several works independently extended the basic ESN architecture by designing more sophisticated reservoirs, readouts or representations of the input.
To evaluate their synergy and efficacy in the context of MTS classification, we introduce a unified framework that generalizes several RC architectures by combining four modules: i) a reservoir module, ii) a dimensionality reduction module, iii) a representation module, and iv) a readout module.
Fig.~\ref{fig:main_schema} gives an overview of the models that can be implemented in the framework (including the proposed reservoir model space), by selecting one option in each module.
The input MTS $\mathbf{X}$ is processed by a reservoir, which is either unidirectional or bidirectional, and it generates over time the states sequence $\mathbf{H}$.
An optional dimensionality reduction step reduces the number of reservoir features and yields a new sequence $\mathbf{\bar{H}}$.
Three different approaches can be chosen to generate the input representation $\mathbf{r}_\mathbf{X}$ from the sequence of reservoir states: the last element in the sequence $\mathbf{h}(T)$, the output state model $\mathbf{\theta}_o$ (Sec.~\ref{sec:esn}), or the proposed reservoir state model $\mathbf{\theta}_h$.
The representation $\mathbf{r}_\mathbf{X}$ is finally processed by a decoder (readout) that predicts the class $\mathbf{y}$.

In the following, we describe the reservoir, dimensionality reduction and readout modules, and we discuss the functionality of the variants implemented in our framework.
A Python software library implementing the unified framework is publicly available online.\footnote{\url{https://github.com/FilippoMB/Reservoir-Computing-framework-for-multivariate-time-series-classification}}

\subsection{Reservoir module}
\label{sec:reservoir}

Several approaches have been proposed to extend the ESN reservoir with additional features, such as the capability of handling multiple time scales~\cite{Gallicchio2017}, or to simplify its large and randomized structure~\cite{rodan_deterministic}.
Of particular interest for the classification of MTS is the bidirectional reservoir, which can replace the standard reservoir in our framework.
RNNs with bidirectional architectures can extract from the input sequence features that account for dependencies very far in time~\cite{graves2005framewise}.
In RC, a bidirectional reservoir has been used in the context of time series prediction to incorporate future information, only provided during training, to improve the accuracy of the model~\cite{rodan2017bidirectional}. 
In a classification setting the whole time series is given at once and, thus, a bidirectional reservoir can be exploited in both training and test to generate better MTS representations~\cite{esn_bidir}.

Bidirectionality is implemented by feeding into the \textit{same} reservoir an input sequence both in straight and reverse order
\begin{equation}
    \label{eq:bidir}
    \begin{aligned}
    \vec{\mathbf{h}}(t) & = f \left( \mathbf{W}_\text{in} \mathbf{x}(t) + \mathbf{W}_\text{r} \vec{\mathbf{h}}(t-1) \right), \\
    \cev{\mathbf{h}}(t) & = f \left( \mathbf{W}_\text{in} \cev{\mathbf{x}}(t) + \mathbf{W}_\text{r} \cev{\mathbf{h}}(t-1) \right),
    \end{aligned}
\end{equation}
where $\cev{\vect{x}}(t) = \mathbf{x}(T-t)$. 
The full state is obtained by concatenating the two state vectors in \eqref{eq:bidir}, and can  capture longer time dependencies by summarizing at every step both recent and past information.

When using a bidirectional reservoir, the linear model in \eqref{eq:res_pred} defining the reservoir model space changes into
\begin{equation}
    \label{eq:bidir_pred}
    \left[ \mathbf{h}(t+1); \cev{\mathbf{h}}(t+1) \right] = \mathbf{U}^b_h  \left[ \vec{\mathbf{h}}(t); \cev{\mathbf{h}}(t) \right] + \mathbf{u}^b_h, 
\end{equation}
where $\mathbf{U}^b_h \in \mathbb{R}^{2R \times 2R}$ and $\mathbf{u}^b_h \in \mathbb{R}^{2R}$ are the new set of parameters.
In this case, the linear model is trained to optimize two distinct objectives: predicting the next state $\mathbf{h}(t+1)$ and reproducing the previous one $\cev{\mathbf{h}}(t+1)$ (or equivalently their low-dimensional embeddings).
We argue that such a model provides a more accurate representation of the input, by modeling temporal dependencies in both time directions to jointly solve a prediction and a memorization task.

\subsection{Dimensionality reduction module}
\label{sec:dim_red}

The dimensionality reduction module projects the sequence of reservoir activations on a lower dimensional subspace, using unsupervised criteria.
In the context of RC, commonly used algorithms for reducing the dimensionality of the reservoir are PCA and kernel PCA, which project data on the first $D$ eigenvectors of a covariance matrix. 
When dealing with a prediction task, dimensionality reduction is applied to a single sequence of reservoir states generated by the input MTS~\cite{lokse2017training}.
On the other hand, in a classification task each MTS is associated to a different sequence of states~\cite{esn_bidir}.
If the MTS are represented by the last reservoir states, those are stacked into a matrix to which standard dimensionality reduction procedures were applied.
When instead the whole set of representations is represented by a tensor, as discussed in Sec.~\ref{sec:reservoir_space}, the dimensionality reduction technique should account for factors of variation across more than one dimension.

Contrarily to the other modules, it is possible to implement a RC classifier without the dimensionality reduction module (as depicted by the skip connection in Fig.~\ref{fig:main_schema}).
However, as discussed in Sec.~\ref{sec:reservoir_space}, dimensionality reduction is particularly important when implementing the proposed reservoir model space representation or when using a bidirectional reservoir, in which cases the size of the representation $\mathbf{r}_\mathbf{X}$ grows with respect to a standard implementation.

\subsection{Readout module}
\label{sec:nonlinear_readouts}

The readout module (decoder) classifies the representations and is either implemented as a linear readout, or a support vector machine (SVM) classifier, or a multi-layer perceptron (MLP).
In a standard ESN, the readout is linear and is quickly trained by solving a convex optimization problem. 
However, a linear readout might not possess sufficient representational power for modeling the embeddings derived from the reservoir states. 
For this reason, several authors proposed to replace the linear decoding function $g(\cdot)$ in \eqref{eq:esn_class} with a nonlinear model, such as SVMs~\cite{bianchi2015prediction} or MLPs~\cite{maass2002real, bush2005modeling, babinec2006merging}. 

Readouts implemented as MLPs accomplished only modest results in the earliest works on RC~\cite{lukovsevivcius2009reservoir}. 
However, nowadays MLPs can be trained much more efficiently by means of sophisticated initialization procedures~\cite{glorot2010understanding} and regularization techniques~\cite{srivastava2014dropout}.
The combination of ESNs with MLPs trained with modern techniques can substantially improve the performance as compared to a linear formulation~\cite{esn_bidir}. 
Following recent trends in the deep learning literature we also investigate endowing the MLP readout with more expressive flexible nonlinear activation functions, namely Maxout~\cite{goodfellow2013maxout} and kernel activation functions~\cite{scardapane2017kafnets}.

\section{Experiments}
\label{sec:experiments}

We test a variety of RC-based architectures for MTS classification implemented with the proposed framework. 
We also compare against RNNs classifiers trained with gradient descent (LSTM and GRU), a $1$\textit{-NN} classifier based on the Dynamic Time Warping (DTW) similarity, SVM classifiers configured with pre-computed kernels for MTS, different deep learning architectures, and other state-of-the-art methods for time series classification.
Depending whether the input MTS in the RC-based model is represented by the last reservoir state ($\mathbf{r}_\vect{X} = \mathbf{h}(T)$), or by the output space model (Sec. \ref{sec:esn}), or by the reservoir space model (Sec. \ref{sec:reservoir_space}), we refer to the models as \textit{lESN}, \textit{omESN} and \textit{rmESN}, respectively.
Whenever we use a bidirectional reservoir, a deep MLP readout, or a SVM readout we add the prefix ``\textit{bi-}'', ``\textit{dr-}'', and ``\textit{svm-}'', respectively (e.g., \textit{bi-lESN} or \textit{dr-bi-rmESN}).

\paragraph{MTS datasets}
To evaluate the performance of each classifier, we consider several MTS classification datasets taken from the UCR\footnote{\url{www.cs.ucr.edu/~eamonn/time_series_data}}, UEA\footnote{\url{https://www.groundai.com/project/the-uea-multivariate-time-series-classification-archive-2018/}}, and UCI repositories\footnote{\url{archive.ics.uci.edu/ml/datasets.html}}. 
For completeness, we also included 3 univariate time series datasets. 
Details of the datasets are reported in Tab.~\ref{tab:benchmark_details}.

\bgroup
\def\arraystretch{0.9} 
\setlength\tabcolsep{.5em} 
\begin{table}[!ht]
\small
\centering
\caption{Time series benchmark datasets details. Column 2 to 5 report the number of variables (\#$V$), samples in training and test set, and number of classes (\#$C$), respectively. 
$T_{min}$ is the length of the shortest MTS in the dataset and $T_{max}$ the longest MTS. All datasets are available at our Github repository.}
\label{tab:benchmark_details}
\begin{tabular}{@{}ll@{ }@{ }l@{ }@{ }@{ }l@{ }@{ }@{ }ll@{ }l@{ }lc@{}}
\cmidrule[1.5pt]{1-8}
\textbf{Dataset} & \#$\boldsymbol{V}$ & \textbf{Train} & \textbf{Test} & \#$\boldsymbol{C}$ & $\boldsymbol{T}_{min}$ & $\boldsymbol{T}_{max}$ & \textbf{Source} \\
\cmidrule[.5pt]{1-8}
Swedish Leaf & 1 & 500 & 625 & 15 & 128 & 128 & UCR \\
Chlorine Conc. & 1 & 467 & 3840 & 3 & 166 & 166 & UCR \\
DistPhal & 1 & 400 & 139 & 3 & 80 & 80 & UCR \\
ECG & 2 & 100 & 100 & 2 & 39 & 152 & UCR \\
Libras & 2 & 180 & 180 & 15 & 45 & 45 & UCI \\
Ch.Traj. & 3 & 300 & 2558 & 20 & 109 & 205 & UCI \\
uWave & 3 & 200 & 427 & 8 & 315 & 315 & UCR \\
NetFlow & 4 & 803 & 534 & 13 & 50 & 994 & UEA \\
Wafer & 6 & 298 & 896 & 2 & 104 & 198 & UCR \\
Robot Fail. & 6 & 100 & 64 & 4 & 15 & 15 & UCI \\
Jp.Vow. & 12 & 270 & 370 & 9 & 7 & 29 & UCI \\
Arab. Dig. & 13 & 6600 & 2200 & 10 & 4 & 93 & UCI \\
Auslan & 22 & 1140 & 1425 & 95 & 45 & 136 & UCI \\
CMUsubject16 & 62 & 29 & 29 & 2 & 127 & 580 & UEA \\
KickvsPunch & 62 & 16 & 10 & 2 & 274 & 841 & UEA \\
WalkvsRun & 62 & 28 & 16 & 2 & 128 & 1918 & UEA \\
PEMS & 963 & 267 & 173 & 7 & 144 & 144 & UCI \\
\cmidrule[1.5pt]{1-8}
\end{tabular}
\end{table}
\egroup

\paragraph{Blood samples dataset}

As a case study on medical data, we analyze MTS of blood measurements obtained from electronic health records of patients undergoing a gastrointestinal surgery at the University Hospital of North Norway in 2004--2012.\footnote{The dataset has been published in the AMIA Data Competition 2016}
Each patient is represented by a MTS of $10$ blood sample measurements collected for $20$ days after surgery.
We consider the problem of classifying patients with and without surgical site infections from their blood samples, collected $20$ days after surgery.
The dataset consists of $883$ MTS, of which $232$ pertain to infected patients.
The original MTS contain missing data, corresponding to measurements not collected for a given patient at certain time intervals, which are replaced by zero-imputation in a preprocessing step.

\paragraph{Experimental setup}
For each dataset, we train the models $10$ times using independent random parameters initializations.
Each model is configured with the same hyperparameters in all the experiments. 
Since reservoirs are sensitive to the hyperparameter setting~\cite{bianchi2017multiplex}, a fine-tuning with independent cross-validation for each task is usually more important in classic RC models than in RNNs trained with gradient descent, such as LSTM and GRU.
Nevertheless, we show that the proposed \textit{rmESN} achieves competitive results even with fixed the hyperparameters.
This indicates higher robustness and gives a practical advantage, compared to traditional RC approaches.

To provide a significant comparison, \textit{lESN}, \textit{omESN} and \textit{rmESN} always share the same randomly generated reservoir configured with the following hyperparameters: number of internal units $R=800$; spectral radius $\rho = 0.99$; non-zero connections percentage $\beta = 0.25$; input scaling $\omega = 0.15$; noise level $\xi = 0.001$.
When classification is performed with a ridge regression readout, we set the regularization value $\lambda = 1.0$.
The ridge regression prediction models, used to generate the model-space representation in \textit{omESN} and \textit{rmESN}, are configured with $\lambda = 5.0$.
We always apply dimensionality reduction, as it provides important computational advantages (both in terms of memory and CPU time), as well as a regularization that improves the generalization capability and robustness of all RC models.
For all experiments we select the number of subspace dimensions as $D=75$, following a grid-search with $k$-fold cross-validation on the datasets of Tab.~\ref{tab:benchmark_details} (see the supplementary material for details).

LSTM and GRU are configured with $H=30$ hidden units; the decoding function is implemented as a neural network with 2 dense layers of $20$ hidden units followed by a softmax layer; the dropout probability is $p_\text{drop} = 0.1$; the $\ell_2$ regularization parameter is $\lambda = 0.0001$; gradient descent is performed with the Adam algorithm~\cite{kingma2014adam} and we train the models for $5000$ epochs
Finally, the 1\textit{-NN} classifier uses FastDTW~\cite{salvador2007toward}, which is a computationally efficient approximation of the DTW\footnote{we used the official Python library: \url{https://pypi.org/project/fastdtw/}}.
We acknowledge additional approaches based on DTW~\cite{gorecki2015multivariate, mei2015learning}, which, however, are not discussed in this paper.

\subsection{Performance comparison on benchmark datasets}
\label{sec:exp1}

\begin{figure}
	\centering
	\includegraphics[keepaspectratio,width=0.9\columnwidth]{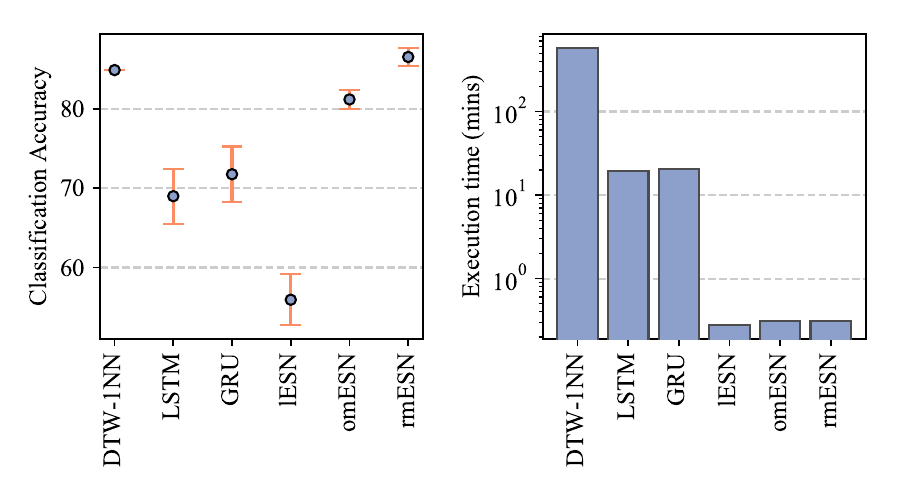}	
    \caption{Comparison of the average results obtained on all benchmark datasets.}
	\label{fig:exp1_res}
\end{figure}

In this experiment we compare the classification accuracy obtained on the representations yielded by the RC models, \textit{lESN}, \textit{omESN} and \textit{rmESN}, by the fully trainable RNNs implementing either GRU or LSTM cells, and by the 1\textbf{-NN} classifier based on DTW.
Evaluation is performed on the benchmark datasets in Tab.~\ref{tab:benchmark_details}.
The decoder is implemented by linear regression in the RC models and by a dense non-linear layer in LSTM and GRU.
Since all the other parameters in LSTM and GRU are learned with gradient descent, the non-linearities in the decoding function do not result in additional computational costs.
Results are reported in Fig.~\ref{fig:exp1_res}.
The first panel reports the mean classification accuracy and standard deviation of 10 independent runs on all benchmark datasets, while the second panel shows the average execution time (in minutes on a logarithmic scale) required for training and testing the models.

The RC classifiers when configured with model space representations achieve a much higher accuracy than the basic \textit{lESN}.
In particular \textit{rmESN}, which adopts our proposed representation, reaches the best overall mean accuracy and the low standard deviation indicates that it is also stable, i.e., it yields consistently good results regardless of the random initialization of the reservoir.
The second-best accuracy is obtained by 1\textit{-NN} with DTW, while the classifiers based on LSTM and GRU perform only better than \textit{lESN}.
The results are particularly interesting since LSTM and GRU exploit supervised information to learn the representations $\mathbf{r}_\vect{X}$ and they adopt a powerful non-linear discriminative classifier.
On the other hand, the RC classifier configured with the model space representation outperforms the other RNN architectures, despite it relies on a linear classifier and the representations are learned in a complete unsupervised fashion. 

In terms of execution time, all the RC classifiers are much faster than the competitors, as the average time for training and test is only few seconds.
Remarkably, thanks to the proposed dimensionality reduction procedure, the \textit{rmESN} classifier can be executed in a time comparable to \textit{lESN}.
The classifiers based on fully trainable RNNs, LSTM and GRU, require in average more than 20 minutes.
Finally, 1\textit{-NN} with DTW is much slower than the other methods despite the adopted ``fast'' implementation~\cite{salvador2007toward}.
This is evident by looking at the huge gap in the execution time, which is more than 11 hours in average and goes beyond 30 hours in some datasets (see the supplementary material for the details).

\subsection{Experiments with bidirectional reservoir and deep-readout}
\label{sec:exp2}

In this experiment we investigate how a bidirectional reservoir and a deep-readout, implemented by a MLP, influence classification accuracy and execution time in the RC-based classifiers.
To further increase the flexibility of the deep readout, beside the standard rectified linear unit (ReLU), we also employ in the MLP more sophisticated transfer functions, namely Maxout~\cite{goodfellow2013maxout} and kernel activation functions (KAFs)~\cite{scardapane2017kafnets}. 
Thanks to their adaptable parameters, trained jointly with the other MLP weights, these functions can improve the expressive capability of the MLP classifier. 
We refer the reader to the original publications for details on their formulation.
The deep readout is implemented with 3 layers of $20$ neurons each and is trained for $5000$ epochs, using a dropout probability $p_\text{drop} = 0.1$ and L\textsubscript{2} regularization parameter $\lambda = 0.001$.

\begin{figure}
	\centering
\includegraphics[keepaspectratio,width=\columnwidth]{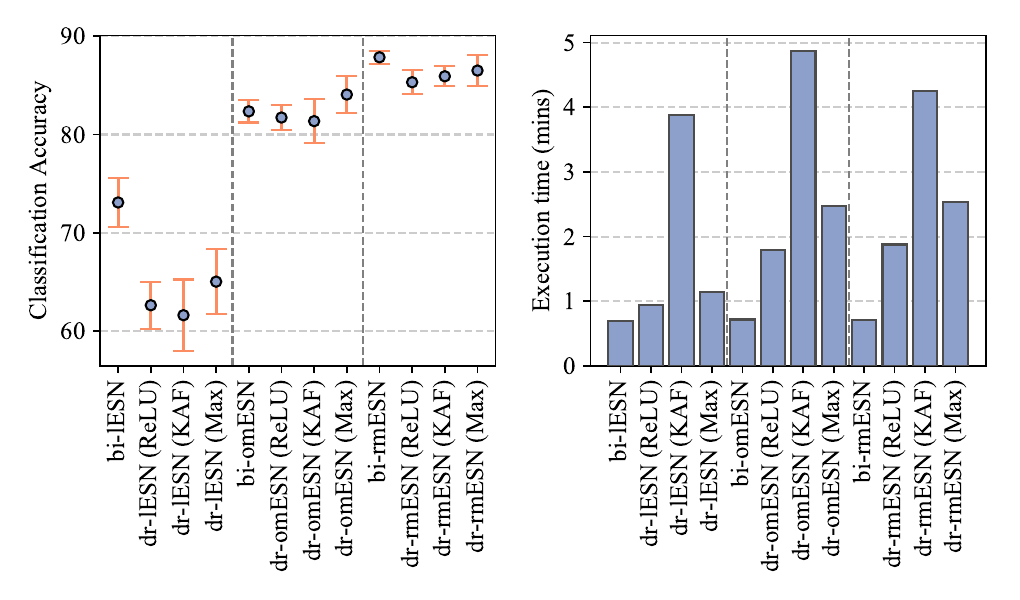}	
    \caption{Classification accuracy and execution time when using RC classifiers with a bidirectional reservoir and deep readouts, configured with ReLUs, KAFs, and Maxout activations.}
	\label{fig:bidir_and_deep}
\end{figure}

We repeat the models evaluation on all the benchmark datasets and in Fig.~\ref{fig:bidir_and_deep} we report results in terms of classification accuracy and training time.
We can see that both the bidirectional reservoir and deep readout improve, to different extents, the classification accuracy of each RC classifier.
The largest improvement occurs for \textit{lESN} when implemented with a bidirectional reservoir. 
This is expected since the last state representation in \textit{lESN} depends mostly on the last observed values of the input MTS.
Whenever the most relevant information is contained at the beginning of the input sequence or when the MTS are too long and the reservoir memory limitation forestall capturing long-term dependencies, the bidirectional architecture greatly improves the \textit{lESN} representation.
The bidirectional reservoir improves the performance also in \textit{omESN} and \textit{rmESN}.
We recall that in these cases, rather than learning only a model for predicting the next output/state, when using a bidirectional reservoir the model also learns to solve a memorization task.
The performance improvement for these model is lower than for \textit{lESN}, probably because the representations obtained with a unidirectional reservoir are already good enough.
Nevertheless, \textit{bi-rmESN} reaches the highest overall accuracy.

A deep-readout enhances the capabilities of the classifier; improvements are larger in \textit{lESN} and more limited in \textit{omESN} and \textit{rmESN}.
Once again, this underlines that the weaker \textit{lESN} representation benefits by adding more complexity in the pipeline.
Even more than the bidirectional reservoir, a deep-readout trades greater modeling capabilities with more computational resources, especially when implemented with adaptive activation functions.
Remarkably, when using Maxout functions rather than a standard ReLU, the training time is slightly higher, but there are significant improvements in the average classification accuracy.
In particular, \textit{dr-omESN (Maxout)} obtains almost the same performance of the basic version of \textit{rmESN}.
Another interesting result obtained by both Maxout and KAF is a reduction in the standard deviation of the accuracy, hence, a more robust classification. 

\begin{figure}
	\centering
	\includegraphics[keepaspectratio,width=0.9\columnwidth]{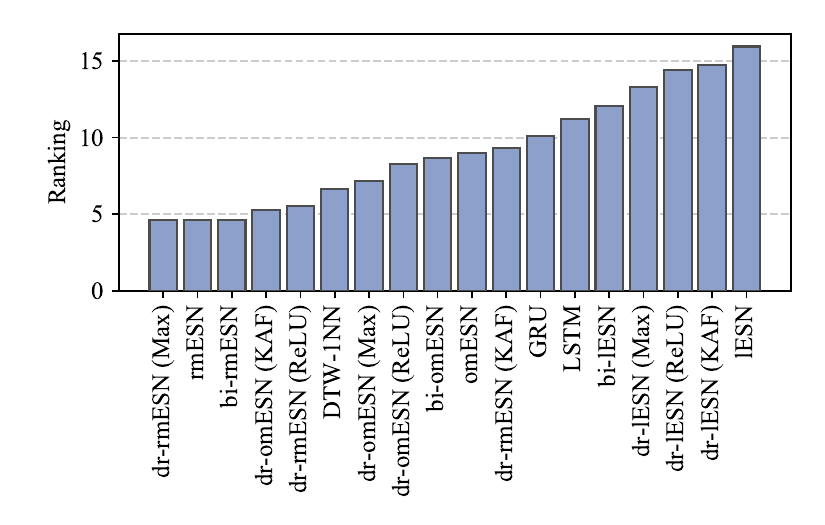}	
    \caption{Ranking in terms of mean accuracy obtained by the MTS classifiers on all the 14 datasets. A lower value in ranking indicates better average accuracy.}
	\label{fig:ranking}
\end{figure}
In Fig.~\ref{fig:ranking} we report the overall ranking, in terms of mean accuracy, of the 18 MTS classifier presented so far on the 17 classification datasets.
On each dataset, the algorithms are ranked from $1$ (best accuracy) to $18$ (worst accuracy) and the table depicts the average of the ranks.
It emerges that the proposed reservoir model space representation is the key factor to achieve the highest classification accuracy and that by introducing further complexity, by means of deep readouts and bidirectional reservoir, performance are further improved.

\subsection{Classification of blood samples MTS}
\label{sec:exp3}

Here, we analyze the blood sample MTS and evaluate the RC classifiers configured with a SVM readout.
We consider only \textit{omESN} and \textit{rmESN} since, as demonstrated in the previous experiments, they provide an optimal compromise between training efficiency and classification accuracy.
Since we adopt a kernel method to implement the decoding function \eqref{eq:rnn_output} (readout), we compare against two state-of-the-art kernels for MTS.
The first is the learned pattern similarity (LPS)~\cite{Baydogan2016}, which identifies segments-occurrence within the MTS by means of regression trees. 
Those are used to generate a bag-of-words type compressed representation, on which the similarity scores are computed.
The second method is the time series cluster kernel (TCK)~\cite{mikalsen2017time}, which is based on an ensemble learning procedure wherein the clustering results of several Gaussian mixture models, which are fit many times on random subsets of the original dataset, are joined to form the final kernel.

For LPS and TCK, an SVM is configured with the pre-computed kernels returned by the two procedures, while for \textit{omESN} and \textit{rmESN} we build a RBF kernel with bandwidth $\gamma$.
We optimize on a validation set the SVM hyperparameters, which are the smoothness of the decision hyperplane, $c$, and bandwidth, $\gamma$ (only \textit{omESN} and \textit{rmESN}).
The hyperparameter space is explored with a grid search, by varying $c$ in $[0.1, 5.0]$ with resolution $0.1$ and $\gamma$ in $[0.01, 1.0]$ with resolution $0.01$.
LPS is configured using 200 regression trees and maximum segments length 10.
TCK is configured with 40 different random initializations and 30 maximum mixtures for each partition.
RC classifiers use the same hyperparameters as in the previous experiments.

\begin{figure}
	\centering
	\includegraphics[keepaspectratio,width=0.9\columnwidth]{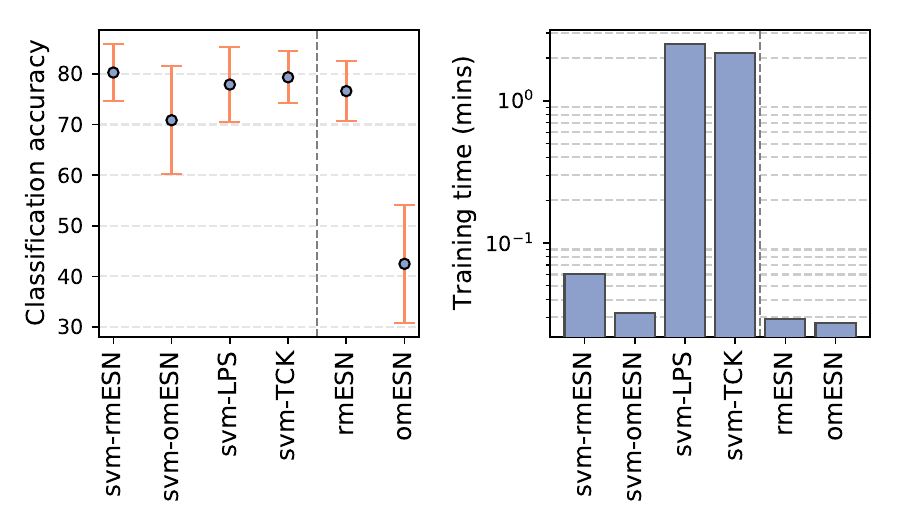}	
    \caption{Classification accuracy obtained with SVM using different precomputed kernels. 
    We also report the results obtained by \textit{rmESN} and \textit{omESN} on the same problem.}
	\label{fig:svm_res}
\end{figure}

To compute the performance of the models, those are evaluated 15 times with independent random initializations and randomly shuffling and splitting the original dataset into training, validation, and test set, containing $70\%$, $10\%$ and $20\%$ of the original samples, respectively.
Each time, we normalize the data by subtracting the mean and dividing by the standard deviation of each variable in the training set, excluding the imputed values. 
The results in terms of classification accuracy and training time are depicted in Fig.~\ref{fig:svm_res}. 
For completeness, we report also the classification results obtained on this task by \textit{omESN} and \textit{rmESN}, with $g(\cdot)$ implemented as a linear readout.
Also in this case, \textit{rmESN} outperforms \textit{omESN} either when it is configured with a linear or a SVM readout.
As for the deep-readout, we notice that the more powerful decoding function improves the classification accuracy in \textit{rmESN} only slightly, while the increment in \textit{omESN} is much larger.
The \textit{svm-rmESN} manages to slightly outperform the SVM classifiers configured with LPS and TCK kernels.
We notice standard deviations in all methods are quite high, since the train/validation/test splits are generated randomly at every iteration and, therefore, the classification task changes each time.
\textit{svm-TCK} yields results with the lowest standard deviation and is followed by \textit{svm-rmESN} and \textit{rmESN}.
The SVM readout slightly increases the training time of the RC models, but they are still much faster than the TCK and LPS kernels.

\bgroup
\def\arraystretch{0.9} 
\setlength\tabcolsep{.5em} 
\begin{table*}[!ht]
\small
\centering
\caption{Results on univariate TS classification. Best results are in bold, second best are underlined.}
\label{tab:univariate_res}
\begin{tabular}{lcccccccccc}
\cmidrule[1.5pt]{1-11}
\textbf{Dataset} & \textbf{MLP} & \textbf{FCN} & \textbf{ResNet} & \textbf{PROP} & \textbf{COTE} & \textbf{BOSS} & \textbf{LSTM-FCN} & \textbf{MMCL} & \textbf{TSML} & \textbf{rm-ESN} \\
\cmidrule[.5pt]{1-11}
\textbf{Adiac} & 24.8 & 14.3 & 17.4 & 35.3 & 23.3 & 30.2 & \textbf{85.9} & 72.6 & 73.7 & \underline{81.2} \\
Chl. Conc. & \textbf{87.2} & 84.3 & 82.8 & 0.64 & 68.6 & 65.5 & 80.1 & -- & -- & \underline{85.6} \\ 
DistPhal & 74.7 & \underline{79.0} & 74.0 & 68.3 & 74.7 & -- & \textbf{81.7} & -- & -- & 75.5 \\
Earthquakes & 79.2 & 80.1 & 78.6 & 71.9 & -- & \underline{80.7} & \textbf{83.5} & -- & -- & 79.7 \\
ECG5000 & 93.5 & 94.1 & 93.1 & 65.0 & -- & 89.0 & \underline{94.7} & -- & -- & \textbf{95.1} \\
FaceAll & 88.5 & 92.9 & 83.4 & 84.8 & 89.5 & 75.9 & \textbf{94.0} & -- & 76.7 & \underline{93.5} \\
FaceFour & 83.0 & 93.2 & 93.2 & 90.1 & 90.1 & \textbf{96.6} & 94.3 & -- & \underline{95.5} & \textbf{96.6} \\
GunPoint & 93.3 & \textbf{100} & \underline{99.3} & \underline{99.3} & \underline{99.3} & \textbf{100} & \textbf{100} & -- & 98 & \textbf{100} \\
ItalyPower & \underline{96.4} & \textbf{97.0} & 96.0 & 96.1 & \underline{96.4} & 91.4 & 96.3 & -- & \underline{96.4} & \underline{96.4} \\
Lightning2 & 72.1 & 80.3 & 75.4 & \textbf{88.5} & \underline{83.6} & 73.8 & 80.3 & 75.4 & 80.3 & 74.2 \\
Swe. Leaf & 89.3 & \underline{96.6} & 95.8 & 91.5 & 95.4 & 85.9 & \textbf{97.9} & -- & 93.0 & 94.5 \\
\cmidrule[1.5pt]{1-11}
\end{tabular}
\end{table*}
\egroup
%

\section{Comparison with deep learning baselines on the classification of univariate time series}
\label{sec:exp4}
Although the proposed framework is specifically designed for the classification of MTS, we conclude by considering additional experiments on univariate time series classification datasets \footnote{we used datasets from \url{http://www.timeseriesclassification.com}}.
Compared to the multivariate case, algorithms designed for this task can exploit stronger biases to attain high classification performance.
We also notice that in the case of univariate time series we do not adopt the proposed extension of PCA for multivariate temporal data, but a regular PCA is used instead.
Nevertheless, we show that our method can achieve competitive results compared to state-of-the-art methods for time series classification. We choose  \textit{rmESN} as the representative model of the RC classifiers, which provides a good trade-off between classification accuracy and training time.
Tab.~\ref{tab:univariate_res} reports the results obtained by \textit{rmESN} and several different methods. We implement baselines based on popular deep learning architectures (MLP, FCN and ResNets)~\cite{fawaz2019deep, wang2017time}, and report results where available on the original papers for BOSS~\cite{schafer2016scalable}, PROP ~\cite{lines2015time}, COTE~\cite{bagnall2015time}, an advanced deep learning architecture that combines an LSTM with attention to a CNN architecture (LSTM-FCN)~\cite{karim2017lstm}, a model-metric co-learning methodology for sequence classification that learns in the model space (MMCL)~\cite{chen2015model}, and a feature-based model (TSML)~\cite{o2017univariate}.

It is possible to see that the complex deep learning architecture LSTM-FCN achieves, on average, the best classification accuracy.
On the other hand, the \textit{rmESN} model equipped with a simple linear readout achieves results that are competitive to those obtained by much more complex models, while requiring only few seconds to be trained.

\section{Conclusions and future work}
\label{sec:conclusions}

We proposed a RC classifier based on the reservoir model space representation, which can be categorized as a hybrid generative-discriminative approach.
Specifically, the parameters of a model that predict the next reservoir states characterize the generative process of the dynamical input features.
Such parameters are, in turn, processed by a discriminative decoder that classifies the original time series.
Usually, in a hybrid generative-discriminative approach where data are assumed to be generated by a parametric distribution, the subsequent discriminative model cannot be specified independently from the generative model type, without introducing biases in the classification~\cite{brodersen2011generative}.  
However, in our case the reservoir is flexible and generic, as it can extract a large variety of
features from the underlying dynamical system, without posing constraints on the particular model underlying the data distribution.
This provides two advantages: (i) different discriminative models can be used in conjunction with the same reservoir model space representation and (ii) the same reservoir can model data from different distributions.

To make the reservoir model space tractable we designed an unsupervised dimensionality reduction procedure, suitable for datasets represented as high-order tensors.
Our dimensionality reduction greatly reduces computational time and memory usage and provides a regularization that prevents overfitting, especially in complex discriminative classifiers.
Finally, we defined a unified framework and investigated several alternatives to build RC classifiers, focusing on unsupervised procedures to learn fixed-size representations of the MTS.

We considered several real-world datasets for classification of MTS, showing that the RC classifier equipped with the proposed representation achieves superior performance both in terms of classification accuracy and execution time. 
We analyzed how a bidirectional reservoir and a deep readout affect the performance (both in time and accuracy) of RC-based classifiers configured with different representations.
We found that combining the reservoir model space with these more sophisticated architectures improves accuracy only slightly, pointing to the already strong informative content of this representation. 
We also considered a medical case study of blood samples time series and obtained superior performance compared to state-of-the-art kernels for MTS.
We concluded by comparing with state-of-the-art methods on the classification of univariate time series and showed that, even on those tasks, our approach achieves competitive results.

\bibliographystyle{IEEEtran}
\bibliography{refs}

\input{supplementary.tex}

\end{document}

%% file: supplementary.tex
\appendix

\subsection{Selection of the optimal number of subspace dimensions}
To determine the optimal number of subspace dimensions $D$, we evaluate how the average training time and classification accuracy (computed with a $k$-fold cross-validation procedure) of the RC classifiers varies on the dataset in Tab.~\ref{tab:benchmark_details}.

\begin{figure}[!ht]
	\centering
	\includegraphics[keepaspectratio,width=0.8\columnwidth]{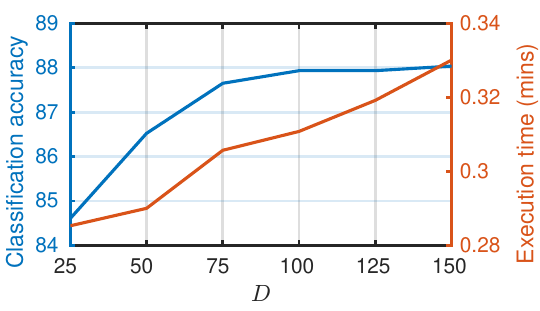}	
    \caption{Average classification accuracy and execution time for different dimensions $D$ of the space with reduced dimensionality.}
	\label{fig:dimred_res}
\end{figure}

We report the average results in Fig. \ref{fig:dimred_res}.

While the training time increases approximately linearly with $D$, it is possible to identify an ``elbow'' in the classification accuracy for $D=75$, which is the value we select in all our experiments.

\subsection{Statistical analysis of the results}
\begin{figure*}[!ht]
	\centering
	\includegraphics[keepaspectratio,width=0.65\textwidth]{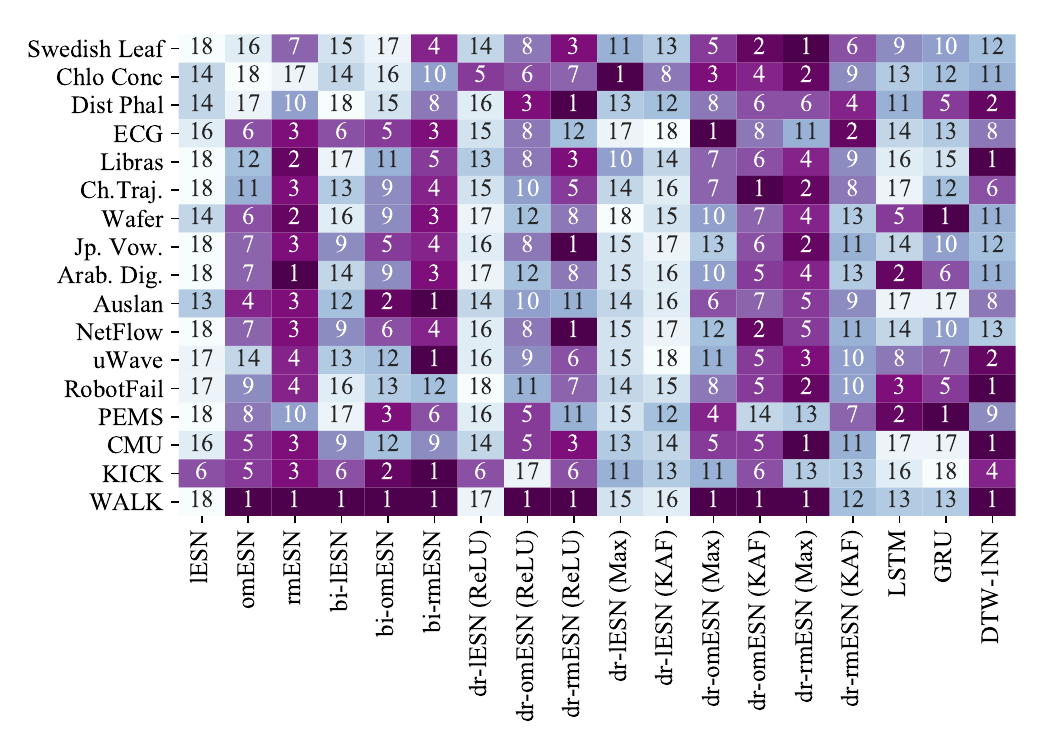}	
    \caption{Ranking of the accuracy obtained by the MTS classifiers on benchmark classification dataset.}
	\label{fig:all_ranks}
\end{figure*}

In the following, we provide the details of the aggregated results reported in Sec.~\ref{sec:exp1} and Sec.~\ref{sec:exp2}. 
Fig.~\ref{fig:all_ranks} depicts the ranking of the accuracy achieved by each MTS classifier on the benchmark datasets described in Tab.~\ref{tab:benchmark_details}. 
Best performance (higher accuracy) correspond to lower values in ranking and to a darker color code.

\begin{figure*}[!ht]
	\centering
	\includegraphics[keepaspectratio,width=0.6\textwidth]{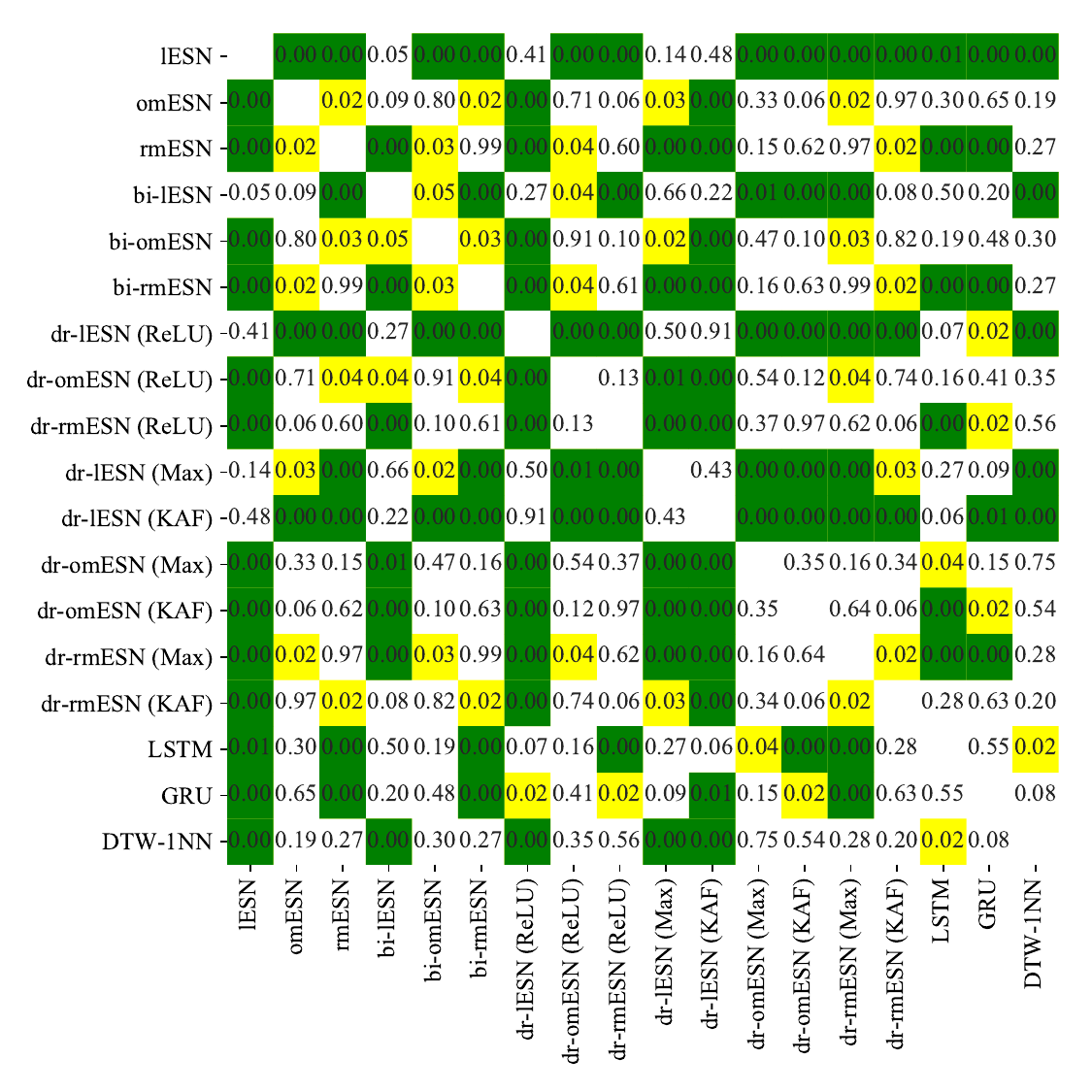}	
    \caption{Results ($p$-values) of the post-hoc test. Yellow boxes indicate $p$-value $< 0.05$, Green boxes indicate $p$-value $< 0.01$.}
	\label{fig:pvals}
\end{figure*}

To evaluate the significance of the differences in performance obtained by the different MTS classifiers on the dataset, we first performed a Friedman test on the rankings. 
We obtained a $p$-value of $1.11e-16$, which indicates the presence of statistically significant differences.
Then, we performed the Finner post-hoc test, to compute for each pair of classifiers if the difference in performance is statistically significant.
In Fig.~\ref{fig:pvals} we report the adjusted $p$-values obtained by testing the performance of each pair of classifiers.
We highlighted in yellow test results with $p$-values lower than $0.05$ and in green the results with $p$-values lower than $0.01$.

\begin{figure*}[!ht]
	\centering
	\includegraphics[keepaspectratio,width=.99\textwidth]{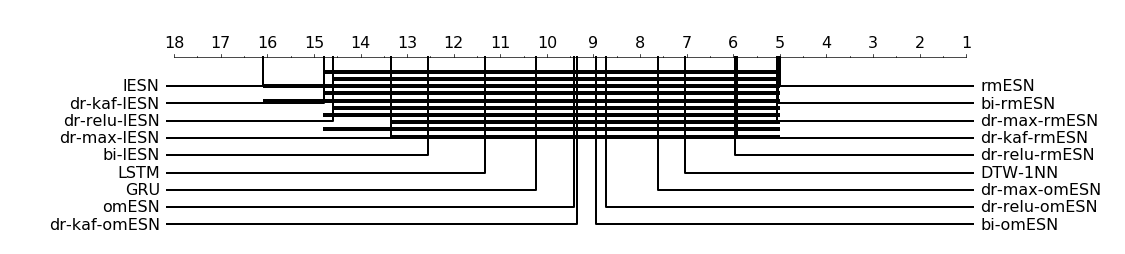}	
    \caption{Critical difference diagram.}
	\label{fig:cd-diag}
\end{figure*}

We also report in Fig.~\ref{fig:cd-diag} a critical-difference diagram based on the Wilcoxon-Holm method to detect pairwise significance. For details about the construction and interpretation of the diagram, we refer the reader to the related Python repository\footnote{\url{https://github.com/hfawaz/cd-diagram}}.

\subsection{Detailed results on the benchmark datasets}
The tables below report the detailed results obtained on each dataset by the time series classifiers analyzed in section Sec.~\ref{sec:exp1} and Sec.~\ref{sec:exp2}.
For each algorithm we performed 10 independent runs and we report the mean accuracy, standard deviation accuracy, mean F1 score, standard deviation F1 score, and mean execution time (in minutes).
For the Arabic Digits dataset we do not report the results for 1-NN with DTW, as the execution time for the simulation exceeded 48 hours.

\begin{table}[!ht]
\footnotesize
\centering
\caption{Results on Swedish Leaf dataset.}
\begin{tabular}{lccc}
\hline
\rowcolor[HTML]{EFEFEF} 
\textbf{\begin{tabular}[c]{@{}l@{}}Swedish Leaf\end{tabular}} 
& \textbf{Accuracy} & \textbf{F1 score} &  \textbf{Time (mins)} \\ \hline
\textit{lESN}           & 62.08$\pm$5.47    & 0.59$\pm$0.06    & 0.15                \\
\textit{omESN}          & 69.47$\pm$2.01    & 0.66$\pm$0.02    & 0.16                \\
\textit{rmESN}          & 94.51$\pm$0.80    & 0.93$\pm$0.01    & 0.16                \\
\textit{bi-lESN}        & 77.98$\pm$4.34    & 0.75$\pm$0.05    & 0.37                \\
\textit{bi-omESN}       & 70.90$\pm$1.81    & 0.68$\pm$0.02    & 0.39                \\
\textit{bi-rmESN}       & 96.25$\pm$0.66    & 0.89$\pm$0.01    & 0.39                \\
\textit{dr-lESN (ReLU)} & 85.14$\pm$0.95    & 0.83$\pm$0.01    & 0.71                \\
\textit{dr-omESN (ReLU)}& 92.79$\pm$2.27    & 0.91$\pm$0.02    & 0.72                \\
\textit{dr-rmESN (ReLU)}& 96.64$\pm$0.77    & 0.94$\pm$0.01    & 1.41                \\
\textit{dr-lESN (Max)}  & 86.42$\pm$1.37    & 0.85$\pm$0.02    & 0.68                \\
\textit{dr-omESN (Max)} & 93.87$\pm$1.25    & 0.92$\pm$0.01    & 0.69                \\
\textit{dr-rmESN (Max)} & 95.56$\pm$0.66    & 0.95$\pm$0.01    & 1.61                \\
\textit{dr-lESN (KAF)}  & 84.43$\pm$2.03    & 0.82$\pm$0.02    & 2.46                \\
\textit{dr-omESN (KAF)} & 92.14$\pm$0.72    & 0.87$\pm$0.01     & 2.51                \\
\textit{dr-rmESN (KAF)} & 95.47$\pm$0.86    & 0.93$\pm$0.01     & 2.58                \\
LSTM                    & 89.58$\pm$0.71    & 0.86$\pm$0.01     & 8.60                \\
GRU                     & 88.24$\pm$1.62    & 0.86$\pm$0.02     & 9.39                \\
DTW-1NN                 & 81.72             & 0.79             & 329.99              \\ \hline
\end{tabular}
\end{table}

\begin{table}[!ht]
\footnotesize
\centering
\caption{Results on Chlorine Concentration dataset.}
\begin{tabular}{lccc}
\hline
\rowcolor[HTML]{EFEFEF} 
\textbf{Chlo Conc}      & \textbf{Accuracy} & \textbf{F1 score} &  \textbf{Time (mins)} \\ \hline
\textit{lESN}              & 68.18$\pm$0.26                   & 0.57$\pm$0.00             & 0.62                \\
\textit{omESN}             & 76.15$\pm$0.28                   & 0.63$\pm$0.01             & 0.68                \\
\textit{rmESN}             & 85.60$\pm$0.41                   & 0.78$\pm$0.01             & 0.70                \\
\textit{bi-lESN}           & 58.18$\pm$0.38                   & 0.49$\pm$0.00             & 1.35                \\
\textit{bi-omESN}          & 77.99$\pm$0.56                   & 0.68$\pm$0.01             & 1.40                \\
\textit{bi-rmESN}          & 83.72$\pm$0.62                   & 0.79$\pm$0.01             & 1.42                \\
\textit{dr-lESN (ReLU)}    & 80.79$\pm$2.09                   & 0.80$\pm$0.02             & 1.10                \\
\textit{dr-omESN (ReLU)}   & 80.38$\pm$0.67                   & 0.80$\pm$0.01             & 1.16                \\
\textit{dr-rmESN (ReLU)}   & 79.68$\pm$0.69                   & 0.79$\pm$0.01             & 1.78                \\
\textit{dr-lESN (Max)}     & 85.95$\pm$1.21                   & 0.86$\pm$0.01             & 1.21                \\
\textit{dr-omESN (Max)}    & 83.05$\pm$0.67                   & 0.83$\pm$0.01             & 1.25                \\
\textit{dr-rmESN (Max)}    & 85.07$\pm$1.36                   & 0.85$\pm$0.01             & 2.14                \\
\textit{dr-lESN (KAF)}     & 72.42$\pm$4.23                   & 0.70$\pm$0.05             & 3.05                \\
\textit{dr-omESN (KAF)}    & 67.04$\pm$2.64                   & 0.66$\pm$0.03             & 3.07                \\
\textit{dr-rmESN (KAF)}    & 81.21$\pm$2.63                   & 0.81$\pm$0.03             & 3.33                \\
LSTM                       & 60.42$\pm$1.10                   & 0.56$\pm$0.03             & 9.07                \\
GRU                        & 60.85$\pm$1.13                   & 0.56$\pm$0.02             & 9.82                \\
DTW-1NN                    & 62.60                            & 0.62                      & 2414.91             \\ \hline
\end{tabular}
\end{table}

\begin{table}[!ht]
\footnotesize
\centering
\caption{Results on Distal Phalanx Outline dataset.}
\begin{tabular}{lccc}
\hline
\rowcolor[HTML]{EFEFEF} 
\textbf{Dist Phal}      & \textbf{Accuracy} & \textbf{F1 score} &  \textbf{Time (mins)} \\ \hline
\textit{lESN}              & 68.92$\pm$0.54                   & 0.67$\pm$0.01             & 0.06                \\
\textit{omESN}             & 67.48$\pm$0.29                   & 0.63$\pm$0.00             & 0.07                \\
\textit{rmESN}             & 75.57$\pm$1.32                   & 0.74$\pm$0.02             & 0.07                \\
\textit{bi-lESN}           & 67.34$\pm$1.08                   & 0.65$\pm$0.01             & 0.20                \\
\textit{bi-omESN}          & 68.06$\pm$0.98                   & 0.64$\pm$0.02             & 0.20                \\
\textit{bi-rmESN}          & 75.23$\pm$0.73                   & 0.72$\pm$0.01             & 0.21                \\
\textit{dr-lESN (ReLU)}    & 67.77$\pm$0.84                   & 0.68$\pm$0.01             & 0.50                \\
\textit{dr-omESN (ReLU)}   & 73.67$\pm$1.55                   & 0.74$\pm$0.02             & 0.50                \\
\textit{dr-rmESN (ReLU)}   & 75.54$\pm$1.02                   & 0.76$\pm$0.01             & 1.03                \\
\textit{dr-lESN (Max)}     & 69.35$\pm$2.35                   & 0.65$\pm$0.02             & 0.62                \\
\textit{dr-omESN (Max)}    & 72.23$\pm$2.97                   & 0.72$\pm$0.03             & 0.62                \\
\textit{dr-rmESN (Max)}    & 76.52$\pm$0.70                   & 0.73$\pm$0.01             & 1.34                \\
\textit{dr-lESN (KAF)}     & 70.22$\pm$1.33                   & 0.70$\pm$0.01             & 2.41                \\
\textit{dr-omESN (KAF)}    & 73.09$\pm$1.91                   & 0.73$\pm$0.02             & 2.39                \\
\textit{dr-rmESN (KAF)}    & 75.52$\pm$2.00                   & 0.72$\pm$0.02             & 2.70                \\
LSTM                       & 70.94$\pm$2.93                   & 0.71$\pm$0.03             & 4.48                \\
GRU                        & 72.66$\pm$1.29                   & 0.73$\pm$0.01             & 4.88                \\
DTW-1NN                    & 74.82                 & 0.75                   & 24.50               \\ \hline
\end{tabular}
\end{table}

\begin{table}[!ht]
\footnotesize
\centering
\caption{Results on Electrocardiography dataset.}
\begin{tabular}{lccc}
\hline
\rowcolor[HTML]{EFEFEF} 
\textbf{ECG}             & \textbf{Accuracy} & \textbf{F1 score} &  \textbf{Time (mins)} \\ \hline
\textit{lESN}                     & 69.00$\pm$2.19                   & 0.81$\pm$0.01             & 0.05                \\
\textit{omESN}             & 84.60$\pm$0.49                   & 0.89$\pm$0.00             & 0.05                \\
\textit{rmESN}           & 85.20$\pm$0.75                   & 0.89$\pm$0.00             & 0.05                \\
\textit{bi-lESN}                  & 84.60$\pm$2.06                   & 0.89$\pm$0.01             & 0.16                \\
\textit{bi-omESN}          & 84.80$\pm$1.17                   & 0.89$\pm$0.01             & 0.16                \\
\textit{bi-rmESN}        & 85.20$\pm$0.40                   & 0.89$\pm$0.00             & 0.17                \\
\textit{dr-lESN (ReLU)}           & 71.60$\pm$4.03                   & 0.79$\pm$0.04             & 0.17                \\
\textit{dr-omESN (ReLU)}   & 84.00$\pm$1.41                   & 0.88$\pm$0.01             & 0.17                \\
\textit{dr-rmESN (ReLU)} & 83.40$\pm$1.62                   & 0.88$\pm$0.01             & 0.28                \\
\textit{dr-lESN (Max)}            & 68.80$\pm$3.43                   & 0.76$\pm$0.03             & 0.20                \\
\textit{dr-omESN (Max)}    & 86.60$\pm$1.02                   & 0.90$\pm$0.01             & 0.21                \\
\textit{dr-rmESN (Max)}  & 83.80$\pm$1.60                   & 0.88$\pm$0.01             & 0.38                \\
\textit{dr-lESN (KAF)}            & 65.60$\pm$7.17                   & 0.74$\pm$0.06             & 0.65                \\
\textit{dr-omESN (KAF)}    & 85.40$\pm$0.49                   & 0.89$\pm$0.00             & 0.64                \\
\textit{dr-rmESN (KAF)}  & 84.00 $\pm$1.10                   & 0.88$\pm$0.01             & 0.69                \\
LSTM                    & 76.20$\pm$4.26                   & 0.82$\pm$0.03             & 2.10                \\
GRU                     & 81.20$\pm$3.49                   & 0.86$\pm$0.02             & 2.27                \\
DTW-1NN                 & 84.00                                  & 0.88                    & 11.42               \\ \hline
\end{tabular}
\end{table}

\begin{table}[!ht]
\footnotesize
\centering
\caption{Results on Libras dataset.}
\begin{tabular}{lccc}
\hline
\rowcolor[HTML]{EFEFEF} 
\textbf{Libras}         & \textbf{Accuracy} & \textbf{F1 score} &  \textbf{Time (mins)} \\ \hline
\textit{lESN}               & 59.89$\pm$0.65                   & 0.59$\pm$0.01             & 0.04                \\
\textit{omESN}              & 77.22$\pm$3.33                   & 0.75$\pm$0.04             & 0.04                \\
\textit{rmESN}              & 88.11$\pm$1.43                   & 0.88$\pm$0.02             & 0.04                \\
\textit{bi-lESN}            & 63.33$\pm$2.30                   & 0.63$\pm$0.02             & 0.13                \\
\textit{bi-omESN}           & 77.78$\pm$0.99                   & 0.77$\pm$0.01             & 0.13                \\
\textit{bi-rmESN}           & 86.00$\pm$0.65                   & 0.86$\pm$0.01             & 0.14                \\
\textit{dr-lESN (ReLU)}     & 72.56$\pm$2.91                   & 0.72$\pm$0.02             & 0.25                \\
\textit{dr-omESN (ReLU)}    & 80.78$\pm$2.29                   & 0.80$\pm$0.02             & 0.26                \\
\textit{dr-rmESN (ReLU)}    & 87.22$\pm$1.76                   & 0.87$\pm$0.02             & 0.48                \\
\textit{dr-lESN (Max)}      & 78.00$\pm$1.43                   & 0.78$\pm$0.02             & 0.29                \\
\textit{dr-omESN (Max)}     & 84.44$\pm$2.17                   & 0.84$\pm$0.02             & 0.30                \\
\textit{dr-rmESN (Max)}     & 86.67$\pm$0.79                   & 0.87$\pm$0.01             & 0.62                \\
\textit{dr-lESN (KAF)}      & 72.22$\pm$2.25                   & 0.72$\pm$0.02             & 1.11                \\
\textit{dr-omESN (KAF)}     & 79.67$\pm$2.40                   & 0.79$\pm$0.02             & 1.11                \\
\textit{dr-rmESN (KAF)}     & 84.78$\pm$1.03                   & 0.85$\pm$0.01             & 1.18                \\
LSTM                        & 68.22$\pm$2.62                   & 0.68$\pm$0.03             & 1.17                \\
GRU                         & 71.56$\pm$4.60                   & 0.71$\pm$0.05             & 1.25                \\
DTW-1NN                     & 88.33                                   & 0.88                    & 9.52                \\ \hline
\end{tabular}
\end{table}

\begin{table}[!ht]
\footnotesize
\centering
\caption{Results on Character Trajectory dataset.}
\begin{tabular}{lccc}
\hline
\rowcolor[HTML]{EFEFEF} 
\textbf{Ch.Traj.}       & \textbf{Accuracy} & \textbf{F1 score} &  \textbf{Time (mins)} \\ \hline
\textit{lESN}               & 21.41$\pm$7.01                   & 0.17$\pm$0.06             & 0.46                \\
\textit{omESN}              & 91.39$\pm$0.91                   & 0.91$\pm$0.01             & 0.50                \\
\textit{rmESN}              & 97.36$\pm$0.24                   & 0.97$\pm$0.00             & 0.51                \\
\textit{bi-lESN}            & 51.11$\pm$8.37                   & 0.49$\pm$0.09             & 1.01                \\
\textit{bi-omESN}           & 94.36$\pm$0.40                   & 0.94$\pm$0.00             & 1.06                \\
\textit{bi-rmESN}           & 97.00$\pm$0.11                   & 0.97$\pm$0.00             & 1.06                \\
\textit{dr-lESN (ReLU)}     & 44.05$\pm$5.12                   & 0.43$\pm$0.05             & 0.82                \\
\textit{dr-omESN (ReLU)}    & 94.08$\pm$0.96                   & 0.94$\pm$0.01             & 0.88                \\
\textit{dr-rmESN (ReLU)}    & 96.58$\pm$0.67                   & 0.97$\pm$0.01             & 1.26                \\
\textit{dr-lESN (Max)}      & 44.71$\pm$4.81                   & 0.44$\pm$0.05             & 0.87                \\
\textit{dr-omESN (Max)}     & 95.54$\pm$0.34                   & 0.95$\pm$0.00             & 0.96                \\
\textit{dr-rmESN (Max)}     & 97.52$\pm$0.54                   & 0.97$\pm$0.01             & 1.47                \\
\textit{dr-lESN (KAF)}      & 40.13$\pm$8.03                   & 0.39$\pm$0.08             & 2.18                \\
\textit{dr-omESN (KAF)}     & 94.50$\pm$0.60                   & 0.94$\pm$0.01             & 2.25                \\
\textit{dr-rmESN (KAF)}     & 97.59$\pm$0.23                   & 0.97$\pm$0.00             & 2.38                \\
LSTM                        & 37.10$\pm$14.62                  & 0.33$\pm$0.16             & 8.50                \\
GRU                         & 70.79$\pm$17.71                  & 0.70$\pm$0.19             & 9.13                \\
DTW-1NN                     & 95.78                                   & 0.96                   & 1218.31             \\ \hline
\end{tabular}
\end{table}

\begin{table}[!ht]
\footnotesize
\centering
\caption{Results on Wafer dataset.}
\begin{tabular}{lccc}
\hline
\rowcolor[HTML]{EFEFEF} 
\textbf{Wafer}          & \textbf{Accuracy} & \textbf{F1 score} &  \textbf{Time (mins)} \\ \hline
\textit{lESN}               & 89.35$\pm$0.09                   & 0.94$\pm$0.00             & 0.22                \\
\textit{omESN}              & 95.71$\pm$1.05                   & 0.98$\pm$0.01             & 0.24                \\
\textit{rmESN}              & 97.78$\pm$0.29                   & 0.98$\pm$0.00             & 0.24                \\
\textit{bi-lESN}            & 88.91$\pm$0.32                   & 0.94$\pm$0.00             & 0.52                \\
\textit{bi-omESN}           & 95.25$\pm$0.78                   & 0.97$\pm$0.00             & 0.54                \\
\textit{bi-rmESN}           & 97.01$\pm$0.40                   & 0.98$\pm$0.00             & 0.54                \\
\textit{dr-lESN (ReLU)}     & 88.50$\pm$0.95                   & 0.94$\pm$0.00             & 0.53                \\
\textit{dr-omESN (ReLU)}    & 94.51$\pm$1.13                   & 0.97$\pm$0.01             & 0.59                \\
\textit{dr-rmESN (ReLU)}    & 95.60$\pm$0.82                   & 0.98$\pm$0.00             & 0.92                \\
\textit{dr-lESN (Max)}      & 88.30$\pm$1.85                   & 0.94$\pm$0.01             & 0.61                \\
\textit{dr-omESN (Max)}     & 95.11$\pm$1.08                   & 0.97$\pm$0.01             & 0.75                \\
\textit{dr-rmESN (Max)}     & 96.85$\pm$0.65                   & 0.98$\pm$0.00             & 1.14                \\
\textit{dr-lESN (KAF)}      & 88.93$\pm$1.45                   & 0.94$\pm$0.01             & 1.85                \\
\textit{dr-omESN (KAF)}     & 93.68$\pm$1.07                   & 0.96$\pm$0.01             & 1.94                \\
\textit{dr-rmESN (KAF)}     & 95.69$\pm$1.00                   & 0.98$\pm$0.01             & 2.02                \\
LSTM                        & 96.32$\pm$3.70                   & 0.98$\pm$0.02             & 7.58                \\
GRU                         & 98.41$\pm$0.86                   & 0.99$\pm$0.00             & 8.22                \\
DTW-1NN                     & 95.09                   & 0.97                  & 396.99              \\ \hline
\end{tabular}
\end{table}

\begin{table}[!ht]
\footnotesize
\centering
\caption{Results on Japanese Vowels dataset.}
\begin{tabular}{lccc}
\hline
\rowcolor[HTML]{EFEFEF} 
\textbf{Jp. Vow.}       & \textbf{Accuracy} & \textbf{F1 score} &  \textbf{Time (mins)} \\ \hline
\textit{lESN}               & 80.00$\pm$5.37                   & 0.80$\pm$0.05             & 0.04                \\
\textit{omESN}              & 95.35$\pm$0.46                   & 0.95$\pm$0.00             & 0.05                \\
\textit{rmESN}              & 97.83$\pm$0.50                   & 0.98$\pm$0.00             & 0.05                \\
\textit{bi-lESN}            & 94.05$\pm$0.70                   & 0.94$\pm$ 0.01            & 0.14                \\
\textit{bi-omESN}           & 97.35$\pm$0.40                   & 0.97$\pm$0.00             & 0.15                \\
\textit{bi-rmESN}           & 97.62$\pm$0.46                   & 0.98$\pm$0.00             & 0.15                \\
\textit{dr-lESN (ReLU)}     & 83.84$\pm$4.25                   & 0.84$\pm$0.04             & 0.32                \\
\textit{dr-omESN (ReLU)}    & 94.76$\pm$0.86                   & 0.95$\pm$0.01             & 0.44                \\
\textit{dr-rmESN (ReLU)}    & 98.14$\pm$0.44                   & 0.97$\pm$0.00             & 0.67                \\
\textit{dr-lESN (Max)}      & 86.22$\pm$3.95                   & 0.86$\pm$0.04             & 0.31                \\
\textit{dr-lESN (KAF)}      & 82.97$\pm$3.90                   & 0.83$\pm$0.04             & 1.18                \\
\textit{dr-omESN (Max)}     & 93.41$\pm$0.40                   & 0.93$\pm$0.00             & 0.46                \\
\textit{dr-omESN (KAF)}     & 93.57$\pm$0.46                   & 0.94$\pm$0.01             & 1.33                \\
\textit{dr-rmESN (KAF)}     & 96.97$\pm$0.63                   & 0.97$\pm$0.01             & 1.24                \\
\textit{dr-rmESN (Max)}     & 97.99$\pm$0.65                   & 0.97$\pm$0.01             & 0.80                \\
LSTM                        & 92.70$\pm$1.36                   & 0.93$\pm$0.01             & 1.15                \\
GRU                         & 94.00$\pm$2.21                   & 0.94$\pm$0.02             & 1.24                \\
DTW-1NN                     & 93.51                                   & 0.94                    & 19.23               \\ \hline
\end{tabular}
\end{table}

\begin{table}[!ht]
\footnotesize
\centering
\caption{Results on Arabic Digits dataset.}
\begin{tabular}{lccc}
\hline
\rowcolor[HTML]{EFEFEF} 
\textbf{Arab. Dig.}     & \textbf{Accuracy} & \textbf{F1 score} &  \textbf{Time (mins)} \\ \hline
\textit{lESN}               & 39.77$\pm$6.08                   & 0.26$\pm$0.06             & 0.92                \\
\textit{omESN}              & 95.63$\pm$0.51                   & 0.95$\pm$0.01             & 1.07                \\
\textit{rmESN}              & 98.12$\pm$0.21                   & 0.98$\pm$0.00             & 1.16                \\
\textit{bi-lESN}            & 77.44$\pm$2.13                   & 0.76$\pm$0.03             & 2.66                \\
\textit{bi-omESN}           & 94.92$\pm$0.27                   & 0.95$\pm$0.00             & 2.80                \\
\textit{bi-rmESN}           & 96.46$\pm$0.44                   & 0.96$\pm$0.00             & 2.90                \\
\textit{dr-lESN (ReLU)}     & 45.82$\pm$2.66                   & 0.45$\pm$0.02             & 6.57                \\
\textit{dr-omESN (ReLU)}    & 92.48$\pm$0.32                   & 0.92$\pm$0.00             & 10.08               \\
\textit{dr-rmESN (ReLU)}    & 95.39$\pm$0.52                   & 0.95$\pm$0.01             & 15.42               \\
\textit{dr-lESN (Max)}      & 46.90$\pm$4.12                   & 0.46$\pm$0.04             & 9.73                \\
\textit{dr-lESN (KAF)}      & 46.11$\pm$3.03                   & 0.45$\pm$0.03             & 40.17               \\
\textit{dr-omESN (Max)}     & 94.01$\pm$0.44                   & 0.94$\pm$0.00             & 16.86               \\
\textit{dr-omESN (KAF)}     & 91.66$\pm$0.59                   & 0.92$\pm$0.01             & 44.52               \\
\textit{dr-rmESN (Max)}     & 96.10$\pm$0.35                   & 0.96$\pm$0.00             & 22.18               \\
\textit{dr-rmESN (KAF)}     & 96.02$\pm$0.76                   & 0.96$\pm$0.01             & 41.16               \\
LSTM                        & 96.61$\pm$0.69                   & 0.97$\pm$0.01             & 82.41               \\
GRU                         & 95.98$\pm$2.91                   & 0.96$\pm$0.03             & 90.82               \\
DTW-1NN                     & --                             & --                     & $>48$ hours             \\ \hline
\end{tabular}
\end{table}

\begin{table}[!ht]
\footnotesize
\centering
\caption{Results on Australian Sign Language dataset.}
\begin{tabular}{lccc}
\hline
\rowcolor[HTML]{EFEFEF} 
\textbf{Auslan}         & \textbf{Accuracy} & \textbf{F1 score} &  \textbf{Time (mins)} \\ \hline
\textit{lESN}               & 1.35$\pm$0.26                    & 0.00$\pm$0.00             & 0.34                \\
\textit{omESN}              & 94.53$\pm$0.43                   & 0.94$\pm$0.00             & 0.39                \\
\textit{rmESN}              & 97.25$\pm$0.25                   & 0.97$\pm$0.00             & 0.40                \\
\textit{bi-lESN}            & 56.94$\pm$0.95                   & 0.56$\pm$0.01             & 0.80                \\
\textit{bi-omESN}           & 97.39$\pm$0.30                   & 0.97$\pm$0.00             & 0.85                \\
\textit{bi-rmESN}           & 97.64$\pm$0.35                   & 0.98$\pm$0.00             & 0.85                \\
\textit{dr-lESN (ReLU)}     & 1.31$\pm$0.28                    & 0.01$\pm$0.00             & 2.09                \\
\textit{dr-omESN (ReLU)}    & 77.40$\pm$2.12                   & 0.77$\pm$0.02             & 3.32                \\
\textit{dr-rmESN (ReLU)}    & 73.47$\pm$4.77                   & 0.73$\pm$0.05             & 4.01                \\
\textit{dr-lESN (Max)}      & 1.31$\pm$0.21                    & 0.01$\pm$0.00             & 2.09                \\
\textit{dr-lESN (KAF)}      & 1.09$\pm$0.08                    & 0.00$\pm$0.00             & 7.65                \\
\textit{dr-omESN (Max)}     & 87.94$\pm$0.44                   & 0.88$\pm$0.00             & 2.66                \\
\textit{dr-rmESN (KAF)}     & 85.75$\pm$0.87                   & 0.86$\pm$0.01             & 8.56                \\
\textit{dr-rmESN (Max)}     & 88.70$\pm$1.38                   & 0.89$\pm$0.01             & 4.49                \\
\textit{dr-omESN (KAF)}     & 84.53$\pm$2.37                   & 0.84$\pm$0.02             & 8.75                \\
LSTM                        & 1.05$\pm$0.00                    & 0.00$\pm$0.00             & 18.89               \\
GRU                         & 1.05$\pm$0.00                    & 0.00$\pm$0.00             & 20.49               \\
DTW-1NN                     & 85.61                                  & 0.85                & 1650.32             \\ \hline
\end{tabular}
\end{table}

\begin{table}[!ht]
\footnotesize
\centering
\caption{Results on Network Flow dataset.}
\begin{tabular}{lccc}
\hline
\rowcolor[HTML]{EFEFEF} 
\textbf{NetFlow}        & \textbf{Accuracy} & \textbf{F1 score} &  \textbf{Time (mins)} \\ \hline
\textit{lESN}               & 79.13$\pm$5.40                   & 0.82$\pm$0.06             & 0.50                \\
\textit{omESN}              & 94.48$\pm$0.50                   & 0.96$\pm$0.01             & 0.51                \\
\textit{rmESN}              & 96.96$\pm$0.54                   & 0.98$\pm$0.01             & 0.52                \\
\textit{bi-lESN}            & 93.19$\pm$0.74                   & 0.96$\pm$0.02             & 1.28                \\
\textit{bi-omESN}           & 95.48$\pm$0.43                   & 0.96$\pm$0.01             & 1.26                \\
\textit{bi-rmESN}           & 96.75$\pm$0.50                   & 0.98$\pm$0.01             & 1.17                \\
\textit{dr-lESN (ReLU)}     & 82.97$\pm$4.29                   & 0.86$\pm$0.05             & 0.88                \\
\textit{dr-omESN (ReLU)}    & 93.89$\pm$0.90                   & 0.97$\pm$0.02             & 0.90                \\
\textit{dr-rmESN (ReLU)}    & 97.27$\pm$0.47                   & 0.98$\pm$0.01             & 1.27                \\
\textit{dr-lESN (Max)}      & 85.35$\pm$3.99                   & 0.88$\pm$0.05             & 0.88                \\
\textit{dr-lESN (KAF)}      & 82.11$\pm$3.93                   & 0.85$\pm$0.05             & 0.98                \\
\textit{dr-omESN (Max)}     & 92.54$\pm$0.44                   & 0.95$\pm$0.01             & 1.54                \\
\textit{dr-omESN (KAF)}     & 92.70$\pm$0.50                   & 0.95$\pm$0.01             & 2.36                \\
\textit{dr-rmESN (Max)}     & 96.11$\pm$0.66                   & 0.98$\pm$0.02             & 2.48                \\
\textit{dr-rmESN (KAF)}     & 97.12$\pm$0.69                   & 0.98$\pm$0.02             & 2.45                \\
LSTM                        & 91.84$\pm$1.39                   & 0.95$\pm$0.02             & 8.72                \\
GRU                         & 93.13$\pm$2.25                   & 0.96$\pm$0.03             & 9.42                \\
DTW-1NN                     & 92.08                                   & 0.94                     & 407.73              \\ \hline
\end{tabular}
\end{table}

\begin{table}[!ht]
\footnotesize
\centering
\caption{Results on uWave dataset.}
\begin{tabular}{lccc}
\hline
\rowcolor[HTML]{EFEFEF} 
\textbf{uWave}          & \textbf{Accuracy} & \textbf{F1 score} &  \textbf{Time (mins)} \\ \hline
\textit{lESN}               & 52.01$\pm$1.53                   & 0.50$\pm$0.02             & 0.42                \\
\textit{omESN}              & 65.42$\pm$1.29                   & 0.64$\pm$0.01             & 0.43                \\
\textit{rmESN}              & 88.88$\pm$0.52                   & 0.89$\pm$0.01             & 0.44                \\
\textit{bi-lESN}            & 66.31$\pm$1.95                   & 0.66$\pm$0.02             & 0.95                \\
\textit{bi-omESN}           & 68.22$\pm$1.28                   & 0.67$\pm$0.01             & 0.97                \\
\textit{bi-rmESN}           & 90.51$\pm$1.16                   & 0.90$\pm$0.01             & 0.97                \\
\textit{dr-lESN (ReLU)}     & 52.48$\pm$1.84                   & 0.51$\pm$0.02             & 0.65                \\
\textit{dr-omESN (ReLU)}    & 71.03$\pm$1.80                   & 0.71$\pm$0.02             & 0.66                \\
\textit{dr-rmESN (ReLU)}    & 84.86$\pm$0.83                   & 0.85$\pm$0.01             & 1.05                \\
\textit{dr-lESN (Max)}      & 53.04$\pm$1.68                   & 0.52$\pm$0.02             & 0.67                \\
\textit{dr-lESN (KAF)}      & 46.73$\pm$1.97                   & 0.46$\pm$0.02             & 0.74                \\
\textit{dr-omESN (Max)}     & 70.47$\pm$2.98                   & 0.70$\pm$0.03             & 0.72                \\
\textit{dr-omESN (KAF)}     & 70.51$\pm$2.24                   & 0.70$\pm$0.02             & 0.76                \\
\textit{dr-rmESN (Max)}     & 89.39$\pm$1.45                   & 0.89$\pm$0.01             & 1.38                \\
\textit{dr-rmESN (KAF)}     & 86.54$\pm$1.48                   & 0.86$\pm$0.02             & 1.16                \\
LSTM                        & 72.52$\pm$1.71                   & 0.72$\pm$0.02             & 21.88               \\
GRU                         & 79.49$\pm$2.65                   & 0.79$\pm$0.03             & 22.99               \\
DTW-1NN                     & 89.46                                    & 0.89               & 189.54              \\ \hline
\end{tabular}
\end{table}

\begin{table}[!ht]
\footnotesize
\centering
\caption{Results on Robotic Arm Failure dataset.}
\begin{tabular}{lccc}
\hline
\rowcolor[HTML]{EFEFEF} 
\textbf{RobotFail}      & \textbf{Accuracy} & \textbf{F1 score} &  \textbf{Time (mins)} \\ \hline
\textit{lESN}               & 50.00$\pm$2.80                   & 0.49$\pm$0.03             & 0.01                \\
\textit{omESN}              & 59.69$\pm$1.82                   & 0.58$\pm$0.02             & 0.01                \\
\textit{rmESN}              & 64.38$\pm$1.17                   & 0.63$\pm$0.01             & 0.01                \\
\textit{bi-lESN}            & 51.56$\pm$2.61                   & 0.51$\pm$0.03             & 0.02                \\
\textit{bi-omESN}           & 55.94$\pm$3.34                   & 0.52$\pm$0.04             & 0.02                \\
\textit{bi-rmESN}           & 56.88$\pm$1.25                   & 0.55$\pm$0.01             & 0.02                \\
\textit{dr-lESN (ReLU)}     & 49.38$\pm$4.15                   & 0.48$\pm$0.04             & 0.12                \\
\textit{dr-omESN (ReLU)}    & 57.50$\pm$3.62                   & 0.56$\pm$0.04             & 0.14                \\
\textit{dr-rmESN (ReLU)}    & 62.81$\pm$1.17                   & 0.61$\pm$0.01             & 0.45                \\
\textit{dr-lESN (Max)}      & 53.75$\pm$2.54                   & 0.52$\pm$0.02             & 0.14                \\
\textit{dr-lESN (KAF)}      & 53.44$\pm$6.20                   & 0.52$\pm$0.06             & 0.18                \\
\textit{dr-omESN (Max)}     & 61.56$\pm$2.12                   & 0.60$\pm$0.02             & 0.18                \\
\textit{dr-omESN (KAF)}     & 57.81$\pm$3.95                   & 0.57$\pm$0.04             & 0.20                \\
\textit{dr-rmESN (Max)}     & 66.25$\pm$1.88                   & 0.64$\pm$0.02             & 0.72                \\
\textit{dr-rmESN (KAF)}     & 63.75$\pm$1.17                   & 0.63$\pm$0.01             & 0.50                \\
LSTM                        & 64.69$\pm$3.22                   & 0.62$\pm$0.03             & 0.67                \\
GRU                         & 63.75$\pm$2.30                   & 0.62$\pm$0.02             & 0.72                \\
DTW-1NN                     & 68.75                                    & 0.67                    & 0.41                \\ \hline
\end{tabular}
\end{table}

\begin{table}[!ht]
\footnotesize
\centering
\caption{Results on Peformance Measurement System.}
\begin{tabular}{lccc}
\hline
\rowcolor[HTML]{EFEFEF} 
\textbf{PEMS}               & \textbf{Accuracy} & \textbf{F1 score} &  \textbf{Time (mins)} \\ \hline
\textit{lESN}               & 49.83$\pm$5.30                   & 0.49$\pm$0.05             & 0.20                \\
\textit{omESN}              & 71.68$\pm$1.51                   & 0.72$\pm$0.01             & 0.30                \\
\textit{rmESN}              & 70.40$\pm$3.79                   & 0.70$\pm$0.04             & 0.21                \\
\textit{bi-lESN}            & 63.70$\pm$2.14                   & 0.63$\pm$0.03             & 0.49                \\
\textit{bi-omESN}           & 73.87$\pm$1.61                   & 0.74$\pm$0.02             & 0.72                \\
\textit{bi-rmESN}           & 72.37$\pm$2.02                   & 0.72$\pm$0.02             & 0.53                \\
\textit{dr-lESN (ReLU)}     & 64.05$\pm$3.13                   & 0.64$\pm$0.03             & 0.50                \\
\textit{dr-omESN (ReLU)}    & 72.49$\pm$1.66                   & 0.73$\pm$0.02             & 9.81                \\
\textit{dr-rmESN (ReLU)}    & 69.48$\pm$3.86                   & 0.69$\pm$0.04             & 1.03                \\
\textit{dr-lESN (Max)}      & 68.55$\pm$1.30                   & 0.68$\pm$0.01             & 0.55                \\
\textit{dr-lESN (KAF)}      & 69.36$\pm$3.08                   & 0.69$\pm$0.03             & 0.64                \\
\textit{dr-omESN (Max)}     & 72.72$\pm$1.12                   & 0.73$\pm$0.01             & 13.99               \\
\textit{dr-omESN (KAF)}     & 71.79$\pm$1.48                   & 0.72$\pm$0.02             & 9.94                \\
\textit{dr-rmESN (Max)}     & 69.02$\pm$3.48                   & 0.69$\pm$0.04             & 1.48                \\
\textit{dr-rmESN (KAF)}     & 68.67$\pm$2.77                   & 0.69$\pm$0.03             & 1.20                \\
LSTM                        & 85.57$\pm$1.57                   & 0.86$\pm$0.02             & 118.64              \\
GRU                         & 89.67$\pm$1.51                   & 0.90$\pm$0.02             & 125.98              \\
DTW-1NN                     & 70.52                                  & 0.70                     & 80.99               \\ \hline
\end{tabular}
\end{table}

\begin{table}
\footnotesize
\centering
\caption{Results on CMUsubject16 dataset.}
\begin{tabular}{lccc}
\hline
\rowcolor[HTML]{EFEFEF} 
\textbf{\begin{tabular}[c]{@{}l@{}}CMU\end{tabular}} & \textbf{Accuracy} & \textbf{F1 score} &  \textbf{Time (mins)} \\ \hline
\textit{lESN}           & 56.89$\pm$8.62 & 0.62$\pm$0.13 & 0.14 \\
\textit{omESN}          & 96.53$\pm$0.01 & 0.96$\pm$0.01 & 0.29 \\
\textit{rmESN}          & 98.27$\pm$1.72 & 0.98$\pm$0.01 & 0.23 \\
\textit{bi-lESN}        & 94.82$\pm$1.72 & 0.95$\pm$0.02 & 0.38 \\
\textit{bi-omESN}       & 89.65$\pm$0.01 & 0.91$\pm$0.01 & 0.37 \\
\textit{bi-rmESN}       & 94.83$\pm$1.72 & 0.95$\pm$0.02 & 0.39 \\
\textit{dr-lESN (ReLU)} & 62.06$\pm$0.12 & 0.66$\pm$0.07 & 0.19 \\
\textit{dr-omESN (ReLU)}& 96.55$\pm$0.01 & 0.96$\pm$0.01 & 0.14 \\
\textit{dr-rmESN (ReLU)}& 98.27$\pm$1.72 & 0.98$\pm$0.01 & 0.30 \\
\textit{dr-lESN (Max)}  & 68.96$\pm$3.44 & 0.74$\pm$0.01 & 0.18 \\
\textit{dr-omESN (Max)} & 96.55$\pm$0.01 & 0.96$\pm$0.01 & 0.23 \\
\textit{dr-rmESN (Max)} & 100.0$\pm$0.00 & 1.00$\pm$0.00 & 3.31 \\
\textit{dr-lESN (KAF)}  & 62.06$\pm$0.13 & 0.67$\pm$0.01 & 0.23 \\
\textit{dr-omESN (KAF)} & 93.11$\pm$3.44 & 0.93$\pm$0.03 & 0.30 \\
\textit{dr-rmESN (KAF)} & 96.55$\pm$0.01 & 0.97$\pm$0.01 & 0.27 \\
LSTM                    & 55.17$\pm$13.79& 0.54$\pm$0.12 & 8.22 \\
GRU                     & 54.79$\pm$9.52 & 0.71$\pm$0.08 & 8.13 \\
DTW-1NN                 & 100.0  & 1.0    & 3.31 \\ 
\hline
\end{tabular}
\end{table}

\vfill\null

\begin{table}
\footnotesize
\centering
\caption{Results on Kick versus Punch dataset.}
\begin{tabular}{lccc}
\hline
\rowcolor[HTML]{EFEFEF} 
\textbf{\begin{tabular}[c]{@{}l@{}}KICK\end{tabular}} & \textbf{Accuracy} & \textbf{F1 score} &  \textbf{Time (mins)} \\ \hline
\textit{lESN}           & 60.37$\pm$0.49 & 0.66$\pm$0.01 & 0.16 \\
\textit{omESN}          & 65.41$\pm$5.39 & 0.58$\pm$0.08 & 0.14 \\
\textit{rmESN}          & 75.02$\pm$4.83 & 0.73$\pm$0.06 & 0.17 \\
\textit{bi-lESN}        & 60.37$\pm$7.34 & 0.67$\pm$0.05 & 0.51 \\
\textit{bi-omESN}       & 85.73$\pm$5.13 & 0.85$\pm$0.05 & 0.33 \\
\textit{bi-rmESN}       & 100.0$\pm$0.00 & 1.0$\pm$0.00 & 0.37 \\
\textit{dr-lESN (ReLU)} & 60.37$\pm$0.55 & 0.74$\pm$0.02 & 0.15 \\
\textit{dr-omESN (ReLU)}& 40.18$\pm$0.08 & 0.51$\pm$0.41 & 0.14 \\
\textit{dr-rmESN (ReLU)}& 60.94$\pm$0.01 & 0.75$\pm$0.01 & 0.11 \\
\textit{dr-lESN (Max)}  & 55.34$\pm$3.94 & 0.36$\pm$0.36 & 0.12 \\
\textit{dr-omESN (Max)} & 55.34$\pm$3.94 & 0.36$\pm$0.36 & 0.17 \\
\textit{dr-rmESN (Max)} & 50.23$\pm$1.98 & 0.37$\pm$0.37 & 0.22 \\
\textit{dr-lESN (KAF)}  & 50.23$\pm$1.98 & 0.37$\pm$0.37 & 0.15 \\
\textit{dr-omESN (KAF)} & 50.66$\pm$1.74 & 0.37$\pm$0.37 & 0.14 \\
\textit{dr-rmESN (KAF)} & 60.50$\pm$0.01 & 0.75$\pm$0.01 & 0.26 \\
LSTM                    & 45.67$\pm$5.46 & 0.33$\pm$0.33 & 0.54 \\
GRU                     & 35.83$\pm$5.97 & 0.38$\pm$0.05 & 0.52 \\
DTW-1NN                 & 70 & 0.66  & 1.08 \\
\hline
\end{tabular}
\end{table}

\begin{table}
\footnotesize
\centering
\caption{Results on Walk versus Run dataset.}
\begin{tabular}{lccc}
\hline
\rowcolor[HTML]{EFEFEF} 
\textbf{\begin{tabular}[c]{@{}l@{}}WALK\end{tabular}} & \textbf{Accuracy} & \textbf{F1 score} &  \textbf{Time (mins)} \\ \hline
\textit{lESN}           & 53.12$\pm$3.12 & 0.69$\pm$0.02 & 0.36 \\
\textit{omESN}          & 100.0$\pm$0.00 & 1.0$\pm$0.00 & 0.33 \\
\textit{rmESN}          & 100.0$\pm$0.00 & 1.0$\pm$0.00 & 0.34 \\
\textit{bi-lESN}        & 100.0$\pm$0.00 & 1.0$\pm$0.00 & 0.90 \\
\textit{bi-omESN}       & 100.0$\pm$0.00 & 1.0$\pm$0.00 & 0.87 \\
\textit{bi-rmESN}       & 100.0$\pm$0.00 & 1.0$\pm$0.00 & 0.83 \\
\textit{dr-lESN (ReLU)} & 59.37$\pm$3.12 & 0.73$\pm$0.03 & 0.42 \\
\textit{dr-omESN (ReLU)}& 100.0$\pm$0.00 & 1.0$\pm$0.00 & 0.43 \\
\textit{dr-rmESN (ReLU)}& 100.0$\pm$0.00 & 1.0$\pm$0.00 & 0.44 \\
\textit{dr-lESN (Max)}  & 71.87$\pm$3.12 & 0.81$\pm$0.01 & 0.37 \\
\textit{dr-omESN (Max)} & 100.0$\pm$0.00 & 1.0$\pm$0.00 & 0.45 \\
\textit{dr-rmESN (Max)} & 100.0$\pm$0.00 & 1.0$\pm$0.00 & 0.48 \\
\textit{dr-lESN (KAF)}  & 65.62$\pm$3.12 & 0.77$\pm$0.02 & 0.40 \\
\textit{dr-omESN (KAF)} & 96.87$\pm$3.12 & 0.98$\pm$0.02 & 0.45 \\
\textit{dr-rmESN (KAF)} & 100.0$\pm$0.00 & 1.0$\pm$0.00 & 0.48 \\
LSTM                    & 75.57$\pm$0.13 & 0.85$\pm$0.07 & 25.28 \\
GRU                     & 75.57$\pm$0.13 & 0.85$\pm$0.07 & 25.58 \\
DTW-1NN                 & 100.0 & 1.0  & 5.48 \\
\hline
\end{tabular}
\end{table}

%% file: main.bbl
\begin{thebibliography}{10}
\providecommand{\url}[1]{#1}
\csname url@samestyle\endcsname
\providecommand{\newblock}{\relax}
\providecommand{\bibinfo}[2]{#2}
\providecommand{\BIBentrySTDinterwordspacing}{\spaceskip=0pt\relax}
\providecommand{\BIBentryALTinterwordstretchfactor}{4}
\providecommand{\BIBentryALTinterwordspacing}{\spaceskip=\fontdimen2\font plus
\BIBentryALTinterwordstretchfactor\fontdimen3\font minus
  \fontdimen4\font\relax}
\providecommand{\BIBforeignlanguage}[2]{{%
\expandafter\ifx\csname l@#1\endcsname\relax
\typeout{** WARNING: IEEEtran.bst: No hyphenation pattern has been}%
\typeout{** loaded for the language `#1'. Using the pattern for}%
\typeout{** the default language instead.}%
\else
\language=\csname l@#1\endcsname
\fi
#2}}
\providecommand{\BIBdecl}{\relax}
\BIBdecl

\bibitem{mikalsen2016learning}
K.~{\O}. Mikalsen, F.~M. Bianchi, C.~Soguero-Ruiz, S.~O. Skr{\o}vseth, R.-O.
  Lindsetmo, A.~Revhaug, and R.~Jenssen, ``Learning similarities between
  irregularly sampled short multivariate time series from {EHRs},'' in
  \emph{Proc. 3rd International Workshop on Pattern Recognition for Healthcare
  Analytics at ICPR 2016}, 2016.

\bibitem{carden2008arma}
E.~Carden and J.~Brownjohn, ``Arma modelled time-series classification for
  structural health monitoring of civil infrastructure,'' \emph{Mechanical
  Systems and Signal Processing}, vol.~22, no.~2, pp. 295--314, 2008.

\bibitem{hunt2015using}
D.~Hunt and D.~Parry, ``Using echo state networks to classify unscripted,
  real-world punctual activity,'' in \emph{Engineering Applications of Neural
  Networks}.\hskip 1em plus 0.5em minus 0.4em\relax Springer, 2015, pp.
  369--378.

\bibitem{trentin2015emotion}
E.~Trentin, S.~Scherer, and F.~Schwenker, ``Emotion recognition from speech
  signals via a probabilistic echo-state network,'' \emph{Pattern Recognition
  Letters}, vol.~66, pp. 4--12, 2015.

\bibitem{salvador2007toward}
S.~Salvador and P.~Chan, ``Toward accurate dynamic time warping in linear time
  and space,'' \emph{Intelligent Data Analysis}, vol.~11, no.~5, pp. 561--580,
  2007.

\bibitem{o2017univariate}
C.~O'Reilly, K.~Moessner, and M.~Nati, ``Univariate and multivariate time
  series manifold learning,'' \emph{Knowledge-Based Systems}, vol. 133, pp.
  1--16, 2017.

\bibitem{bagnall2017great}
A.~Bagnall, J.~Lines, A.~Bostrom, J.~Large, and E.~Keogh, ``The great time
  series classification bake off: a review and experimental evaluation of
  recent algorithmic advances,'' \emph{Data Mining and Knowledge Discovery},
  vol.~31, no.~3, pp. 606--660, 2017.

\bibitem{baydogan2015learning}
M.~G. Baydogan and G.~Runger, ``Learning a symbolic representation for
  multivariate time series classification,'' \emph{Data Mining and Knowledge
  Discovery}, vol.~29, no.~2, pp. 400--422, 2015.

\bibitem{graves2013speechJ}
A.~Graves, A.-r. Mohamed, and G.~Hinton, ``Speech recognition with deep
  recurrent neural networks,'' in \emph{Proc. 2013 IEEE International
  Conference on Acoustics, Speech and Signal Processing (ICASSP),}.\hskip 1em
  plus 0.5em minus 0.4em\relax IEEE, 2013, pp. 6645--6649.

\bibitem{lukovsevivcius2009reservoir}
M.~Luko{\v{s}}evi{\v{c}}ius and H.~Jaeger, ``Reservoir computing approaches to
  recurrent neural network training,'' \emph{Computer Science Review}, vol.~3,
  no.~3, pp. 127--149, 2009.

\bibitem{scardapane2017randomness}
S.~Scardapane and D.~Wang, ``Randomness in neural networks: an overview,''
  \emph{Wiley Interdisciplinary Reviews: Data Mining and Knowledge Discovery},
  vol.~7, no.~2, 2017.

\bibitem{bianchi2015prediction}
F.~Bianchi, S.~Scardapane, A.~Uncini, A.~Rizzi, and A.~Sadeghian, ``Prediction
  of telephone calls load using {E}cho {S}tate {N}etwork with exogenous
  variables,'' \emph{Neural Networks}, vol.~71, pp. 204--213, 2015.

\bibitem{7286732}
F.~M. Bianchi, E.~De~Santis, A.~Rizzi, and A.~Sadeghian, ``Short-term electric
  load forecasting using echo state networks and {PCA} decomposition,''
  \emph{IEEE Access}, vol.~3, pp. 1931--1943, Oct. 2015.

\bibitem{rodan2017bidirectional}
A.~Rodan, A.~Sheta, and H.~Faris, ``Bidirectional reservoir networks trained
  using {SVM}$+$ privileged information for manufacturing process modeling,''
  \emph{Soft Computing}, vol.~21, no.~22, pp. 6811--6824, 2017.

\bibitem{jaeger2001echo}
H.~Jaeger, ``The ``echo state'' approach to analysing and training recurrent
  neural networks-with an erratum note,'' \emph{GMD Technical Report}, vol.
  148, no.~34, 2001.

\bibitem{ma2016functional}
Q.~Ma, L.~Shen, W.~Chen, J.~Wang, J.~Wei, and Z.~Yu, ``Functional echo state
  network for time series classification,'' \emph{Information Sciences}, vol.
  373, pp. 1--20, 2016.

\bibitem{skowronski2006minimum}
M.~Skowronski and J.~Harris, ``Minimum mean squared error time series
  classification using an echo state network prediction model,'' in \emph{Proc.
  2006 IEEE International Symposium on Circuits and Systems (ISCAS)}.\hskip 1em
  plus 0.5em minus 0.4em\relax IEEE, 2006.

\bibitem{aswolinskiy2016time}
W.~Aswolinskiy, R.~Reinhart, and J.~Steil, ``Time series classification in
  reservoir-and model-space: a comparison,'' in \emph{IAPR Workshop on
  Artificial Neural Networks in Pattern Recognition}.\hskip 1em plus 0.5em
  minus 0.4em\relax Springer, 2016, pp. 197--208.

\bibitem{chen2013model}
H.~Chen, F.~Tang, P.~Tino, and X.~Yao, ``Model-based kernel for efficient time
  series analysis,'' in \emph{Proceedings of the 19th ACM SIGKDD international
  conference on Knowledge discovery and data mining}.\hskip 1em plus 0.5em
  minus 0.4em\relax ACM, 2013, pp. 392--400.

\bibitem{chen2014learning}
H.~Chen, P.~Tino, A.~Rodan, and X.~Yao, ``Learning in the model space for
  cognitive fault diagnosis,'' \emph{IEEE Transactions on Neural Networks and
  Learning Systems}, vol.~25, no.~1, pp. 124--136, 2014.

\bibitem{lokse2017training}
S.~L{\o}kse, F.~M. Bianchi, and R.~Jenssen, ``Training echo state networks with
  regularization through dimensionality reduction,'' \emph{Cognitive
  Computation}, vol.~9, no.~3, pp. 364--378, Jun 2017.

\bibitem{bianchi2017recurrent}
F.~M. Bianchi, E.~Maiorino, M.~C. Kampffmeyer, A.~Rizzi, and R.~Jenssen,
  \emph{Recurrent Neural Networks for Short-Term Load Forecasting: An Overview
  and Comparative Analysis}.\hskip 1em plus 0.5em minus 0.4em\relax Springer,
  2017.

\bibitem{sutskever2014sequence}
I.~Sutskever, O.~Vinyals, and Q.~V. Le, ``Sequence to sequence learning with
  neural networks,'' in \emph{Advances in neural information processing
  systems}, 2014, pp. 3104--3112.

\bibitem{srivastava2014dropout}
N.~Srivastava, G.~Hinton, A.~Krizhevsky, I.~Sutskever, and R.~Salakhutdinov,
  ``Dropout: A simple way to prevent neural networks from overfitting,''
  \emph{The Journal of Machine Learning Research}, vol.~15, no.~1, pp.
  1929--1958, 2014.

\bibitem{zaremba2014recurrent}
W.~Zaremba, I.~Sutskever, and O.~Vinyals, ``Recurrent neural network
  regularization,'' \emph{arXiv preprint arXiv:1409.2329}, 2014.

\bibitem{pascanu2013difficulty}
R.~Pascanu, T.~Mikolov, and Y.~Bengio, ``On the difficulty of training
  recurrent neural networks,'' in \emph{International Conference on Machine
  Learning}, 2013, pp. 1310--1318.

\bibitem{hochreiter1997long}
S.~Hochreiter and J.~Schmidhuber, ``Long short-term memory,'' \emph{Neural
  computation}, vol.~9, no.~8, pp. 1735--1780, 1997.

\bibitem{cho2014learning}
K.~Cho, B.~Van~Merri{\"e}nboer, C.~Gulcehre, D.~Bahdanau, F.~Bougares,
  H.~Schwenk, and Y.~Bengio, ``Learning phrase representations using rnn
  encoder-decoder for statistical machine translation,'' \emph{arXiv preprint
  arXiv:1406.1078}, 2014.

\bibitem{7765110}
F.~M. Bianchi, L.~Livi, and C.~Alippi, ``Investigating echo-state networks
  dynamics by means of recurrence analysis,'' \emph{IEEE Transactions on Neural
  Networks and Learning Systems}, vol.~29, no.~2, pp. 427--439, Feb 2018.

\bibitem{livi2018determination}
L.~Livi, F.~M. Bianchi, and C.~Alippi, ``Determination of the edge of
  criticality in echo state networks through fisher information maximization,''
  \emph{IEEE transactions on neural networks and learning systems}, vol.~29,
  no.~3, pp. 706--717, 2018.

\bibitem{gong2018multiobjective}
Z.~Gong, H.~Chen, B.~Yuan, and X.~Yao, ``Multiobjective learning in the model
  space for time series classification,'' \emph{IEEE transactions on
  cybernetics}, vol.~49, no.~3, pp. 918--932, 2018.

\bibitem{wang2016effective}
L.~Wang, Z.~Wang, and S.~Liu, ``An effective multivariate time series
  classification approach using echo state network and adaptive differential
  evolution algorithm,'' \emph{Expert Systems with Applications}, vol.~43, pp.
  237--249, 2016.

\bibitem{oord2018representation}
A.~v.~d. Oord, Y.~Li, and O.~Vinyals, ``Representation learning with
  contrastive predictive coding,'' \emph{arXiv preprint arXiv:1807.03748},
  2018.

\bibitem{ng2002discriminative}
A.~Y. Ng and M.~I. Jordan, ``On discriminative vs. generative classifiers: A
  comparison of logistic regression and naive bayes,'' in \emph{Advances in
  neural information processing systems}, 2002, pp. 841--848.

\bibitem{esn_bidir}
F.~M. Bianchi, S.~Scardapane, S.~L{\o}kse, and R.~Jenssen, ``Bidirectional
  deep-readout echo state networks,'' in \emph{European Symposium on Artificial
  Neural Networks}, 2018.

\bibitem{BIANCHI2019106973}
F.~M. Bianchi, L.~Livi, K.~{\O}. Mikalsen, M.~Kampffmeyer, and R.~Jenssen,
  ``Learning representations of multivariate time series with missing data,''
  \emph{Pattern Recognition}, vol.~96, p. 106973, 2019.

\bibitem{kolda2009tensor}
T.~G. Kolda and B.~W. Bader, ``Tensor decompositions and applications,''
  \emph{SIAM review}, vol.~51, no.~3, pp. 455--500, 2009.

\bibitem{Gallicchio2017}
C.~Gallicchio and A.~Micheli, ``Echo state property of deep reservoir computing
  networks,'' \emph{Cognitive Computation}, vol.~9, no.~3, pp. 337--350, Jun
  2017.

\bibitem{rodan_deterministic}
A.~Rodan and P.~Tino, ``Simple deterministically constructed cycle reservoirs
  with regular jumps,'' \emph{Neural Computation}, vol.~24, no.~7, pp.
  1822--1852, 2012, pMID: 22428595.

\bibitem{graves2005framewise}
A.~Graves and J.~Schmidhuber, ``Framewise phoneme classification with
  bidirectional {LSTM} and other neural network architectures,'' \emph{Neural
  Networks}, vol.~18, no. 5-6, pp. 602--610, 2005.

\bibitem{maass2002real}
W.~Maass, T.~Natschl{\"a}ger, and H.~Markram, ``Real-time computing without
  stable states: A new framework for neural computation based on
  perturbations,'' \emph{Neural computation}, vol.~14, no.~11, pp. 2531--2560,
  2002.

\bibitem{bush2005modeling}
K.~Bush and C.~Anderson, ``Modeling reward functions for incomplete state
  representations via echo state networks,'' in \emph{Proc. International Joint
  Conference on Neural Networks (IJCNN)}, vol.~5.\hskip 1em plus 0.5em minus
  0.4em\relax IEEE, 2005, pp. 2995--3000.

\bibitem{babinec2006merging}
{\v{S}}.~Babinec and J.~Posp{\'\i}chal, ``Merging echo state and feedforward
  neural networks for time series forecasting,'' in \emph{International
  Conference on Artificial Neural Networks}.\hskip 1em plus 0.5em minus
  0.4em\relax Springer, 2006, pp. 367--375.

\bibitem{glorot2010understanding}
X.~Glorot and Y.~Bengio, ``Understanding the difficulty of training deep
  feedforward neural networks,'' in \emph{Proceedings of the Thirteenth
  International Conference on Artificial Intelligence and Statistics}, 2010,
  pp. 249--256.

\bibitem{goodfellow2013maxout}
I.~J. Goodfellow, D.~Warde-Farley, M.~Mirza, A.~C. Courville, and Y.~Bengio,
  ``Maxout networks.'' \emph{Proc. 30th International Conference on Machine
  Learning (ICML)}, 2013.

\bibitem{scardapane2017kafnets}
S.~Scardapane, S.~Van~Vaerenbergh, S.~Totaro, and A.~Uncini, ``Kafnets:
  kernel-based non-parametric activation functions for neural networks,''
  \emph{arXiv preprint arXiv:1707.04035}, 2017.

\bibitem{bianchi2017multiplex}
F.~M. Bianchi, L.~Livi, C.~Alippi, and R.~Jenssen, ``Multiplex visibility
  graphs to investigate recurrent neural network dynamics,'' \emph{Scientific
  reports}, vol.~7, p. 44037, 2017.

\bibitem{kingma2014adam}
D.~P. Kingma and J.~Ba, ``Adam: A method for stochastic optimization,''
  \emph{arXiv preprint arXiv:1412.6980}, 2014.

\bibitem{gorecki2015multivariate}
T.~G{\'o}recki and M.~{\L}uczak, ``Multivariate time series classification with
  parametric derivative dynamic time warping,'' \emph{Expert Systems with
  Applications}, vol.~42, no.~5, pp. 2305--2312, 2015.

\bibitem{mei2015learning}
J.~Mei, M.~Liu, Y.-F. Wang, and H.~Gao, ``Learning a mahalanobis distance-based
  dynamic time warping measure for multivariate time series classification,''
  \emph{IEEE transactions on Cybernetics}, vol.~46, no.~6, pp. 1363--1374,
  2015.

\bibitem{Baydogan2016}
M.~G. Baydogan and G.~Runger, ``Time series representation and similarity based
  on local autopatterns,'' \emph{Data Mining and Knowledge Discovery}, vol.~30,
  no.~2, pp. 476--509, 2016.

\bibitem{mikalsen2017time}
K.~{\O}. Mikalsen, F.~M. Bianchi, C.~Soguero-Ruiz, and R.~Jenssen, ``Time
  series cluster kernel for learning similarities between multivariate time
  series with missing data,'' \emph{Pattern Recognition}, vol.~76, pp. 569 --
  581, 2018.

\bibitem{fawaz2019deep}
H.~I. Fawaz, G.~Forestier, J.~Weber, L.~Idoumghar, and P.-A. Muller, ``Deep
  learning for time series classification: a review,'' \emph{Data Mining and
  Knowledge Discovery}, vol.~33, no.~4, pp. 917--963, 2019.

\bibitem{wang2017time}
Z.~Wang, W.~Yan, and T.~Oates, ``Time series classification from scratch with
  deep neural networks: A strong baseline,'' in \emph{2017 international joint
  conference on neural networks (IJCNN)}.\hskip 1em plus 0.5em minus
  0.4em\relax IEEE, 2017, pp. 1578--1585.

\bibitem{schafer2016scalable}
P.~Sch{\"a}fer, ``Scalable time series classification,'' \emph{Data Mining and
  Knowledge Discovery}, vol.~30, no.~5, pp. 1273--1298, 2016.

\bibitem{lines2015time}
J.~Lines and A.~Bagnall, ``Time series classification with ensembles of elastic
  distance measures,'' \emph{Data Mining and Knowledge Discovery}, vol.~29,
  no.~3, pp. 565--592, 2015.

\bibitem{bagnall2015time}
A.~Bagnall, J.~Lines, J.~Hills, and A.~Bostrom, ``Time-series classification
  with cote: the collective of transformation-based ensembles,'' \emph{IEEE
  Transactions on Knowledge and Data Engineering}, vol.~27, no.~9, pp.
  2522--2535, 2015.

\bibitem{karim2017lstm}
F.~Karim, S.~Majumdar, H.~Darabi, and S.~Chen, ``Lstm fully convolutional
  networks for time series classification,'' \emph{IEEE Access}, vol.~6, pp.
  1662--1669, 2017.

\bibitem{chen2015model}
H.~Chen, F.~Tang, P.~Tino, A.~G. Cohn, and X.~Yao, ``Model metric co-learning
  for time series classification,'' in \emph{Twenty-Fourth International Joint
  Conference on Artificial Intelligence}, 2015.

\bibitem{brodersen2011generative}
K.~H. Brodersen, T.~M. Schofield, A.~P. Leff, C.~S. Ong, E.~I. Lomakina, J.~M.
  Buhmann, and K.~E. Stephan, ``Generative embedding for model-based
  classification of fmri data,'' \emph{PLoS computational biology}, vol.~7,
  no.~6, p. e1002079, 2011.

\end{thebibliography}
